%% file: main.tex
\documentclass{article}

\usepackage[colorlinks,linkcolor=black,citecolor=blue,urlcolor=blue,bookmarks=true]{hyperref}
\usepackage{amsfonts}
\usepackage{stackengine}
\usepackage{amsmath}
\usepackage{amssymb}
\usepackage{chngcntr}
\usepackage{graphicx}
\usepackage{bm}
\usepackage{caption}
\usepackage{subcaption}
\usepackage{booktabs}
\usepackage{xcolor}
\usepackage{multirow}
\usepackage{algorithm}
\usepackage{algorithmic}
\usepackage{wrapfig}
\usepackage{tablefootnote}
\usepackage{url}
\newcommand{\TC}[1]{#1}

\usepackage{amsthm}
\newtheorem{definition}{Definition}[section]
\newtheorem{theorem}{Theorem}[section]
\newtheorem*{theorem_r}{Theorem}

\newcommand{\EXP}[2]{\mathbb{E}_{#1}\left[#2\right]}

\newcommand{\params}{\mathbf{\theta}}
\def \vx {\mathbf{x}}   
\def \vy {\mathbf{y}}   
\def \mx {\mathbf{X}}  
\def \my {\mathbf{Y}}  
\def \ma {\mathbf{A}}  
\def \mb {\mathbf{B}}  
\def \muhat {\hat{\mu}} 
\def \nuhat {\hat{\nu}}

\usepackage[nonatbib,final]{neurips_2021}

\title{Don't Generate Me:\\ Training Differentially Private Generative Models with Sinkhorn Divergence}

\author{%
  Tianshi Cao$^{1,2,4}$
  \hspace{0.5cm}
  Alex Bie$^{3}$\thanks{Work done during internship at NVIDIA.}
  \hspace{0.5cm}
  Arash Vahdat$^{4}$
  \hspace{0.5cm}
  Sanja Fidler$^{1,2,4}$
  \hspace{0.5cm}
  Karsten Kreis$^{4}$
  \vspace{2pt}\\
  \small{\textsuperscript{1}University of Toronto \quad \textsuperscript{2}Vector Institute \quad \textsuperscript{3}University of Waterloo \quad \textsuperscript{4}NVIDIA}  \vspace{2pt}\\
  \texttt{\scriptsize
  tianshic@nvidia.com, \quad
  yabie@uwaterloo.ca, \quad
  \{avahdat,sfidler,kkreis\}@nvidia.com
  }
}

\begin{document}

\maketitle

\begin{abstract}
Although machine learning models trained on massive data have led to breakthroughs in several areas, their deployment in privacy-sensitive domains remains limited due to restricted access to data. Generative models trained with privacy constraints on private data can sidestep this challenge, providing indirect access to private data instead. We propose DP-Sinkhorn, a novel optimal transport-based generative method for learning data distributions from private data with differential privacy. DP-Sinkhorn minimizes
the Sinkhorn divergence, a computationally efficient approximation to the exact optimal transport distance, between the model and data in a differentially private manner and 
uses a novel technique for controlling the bias-variance trade-off of gradient estimates.
Unlike existing approaches for training differentially private generative models, which are mostly based on generative adversarial networks, we do not rely on adversarial objectives, which are notoriously difficult to optimize, especially in the presence of noise imposed by privacy constraints. Hence, DP-Sinkhorn is easy to train and deploy. Experimentally, 
we improve upon the state-of-the-art on multiple image modeling benchmarks and show differentially private synthesis of informative RGB images. 
Project page: \url{https://nv-tlabs.github.io/DP-Sinkhorn}.
\end{abstract}

\section{Introduction}
\vspace{-6pt}
Modern machine learning (ML) algorithms and their practical applications (e.g. recommender systems~\cite{gomezuribe2016recsys}, personalized medicine~\cite{ho2020medicine}, face recognition~\cite{wang2020deep}, speech synthesis~\cite{oord2016wavenet}, etc.) have become increasingly data hungry and the use of personal data is often a necessity. Consequently, the importance of privacy protection has become apparent to both the public and academia.

Differential privacy (DP) is a rigorous definition of privacy that quantifies the amount of information leaked by a user, participating in a data release~\cite{dwork2006calibrating,dwork2014diffprivacy}. The degree of privacy protection is represented by the privacy budget.
DP was originally designed for answering queries to statistical databases. In a typical setting, a data analyst (party wanting to use data; e.g. a healthcare
company) sends a query to a data curator (party in charge of safekeeping the database; e.g. a hospital), who makes the query on the database and replies with a semi-random answer that preserves privacy. Responding to each new query incurs a privacy cost. If the analyst has multiple queries, the curator must subdivide the privacy budget to spend on each query. Once the budget is depleted, the curator can no longer respond to queries, preventing the analyst from performing new, unanticipated tasks with the database.

Generative models can be applied as a general and flexible data-sharing medium~\cite{xie2018dpgan,Augenstein2020Generative}, sidestepping the above problems. 
In this scenario, the curator first encodes private data into a generative model; then, the model is shared with the analyst, who can use it to synthesize similar yet different data from the training data. This data can be used in any way desired, 
such as for 
data analysis or to train specific ML models. Unanticipated novel tasks can be accommodated without repeatedly interacting with the curator, since the analyst can easily generate additional synthetic data as required.

Furthermore, it has been observed that generative models can reveal critical information about their training data \cite{webster2021person,hayes2019logan}. For example, Webster et al.~\cite{webster2021person} found that modern GANs trained on images of faces produce examples that greatly resemble their training data, thereby leaking private information. Hence, the generative model must be learnt with privacy constraints to protect the privacy of individuals contributing to the database.

Differentially private learning of generative models has been studied mostly using generative adversarial networks (GANs)~\cite{xie2018dpgan,frigerio2019dpgan,yoon2018pategan,chen2020dpwgan,wang2021datalens}. While GANs in the non-private setting 
can
synthesize complex data such as high definition images~\cite{brock2018biggan,karras2020analyzing}, their application in the private setting is challenging. This is in part because GANs suffer from training instabilities~\cite{arjovsky2017towards,mescheder2018ganconvergence}, which can be exacerbated by adding noise to the GAN's gradients during training, a common technique to implement DP. 
Hence, GANs typically require careful hyperparameter tuning.
This goes against the principle of privacy, where repeated access to data need to be avoided \cite{NIPS2013_5014}. 

In this paper, we propose \textit{DP-Sinkhorn}, a novel method to train differentially private generative models using a semi-debiased Sinkhorn loss. 
DP-Sinkhorn is based on the framework of optimal transport (OT), where the problem of learning a generative model is framed as minimizing the optimal transport distance, a type of Wasserstein distance, between the generator-induced distribution and the real data distribution~\cite{bousquet2017vegan,peyre2019ot}.
DP-Sinkhorn approximates the exact OT distance in the primal space using the Sinkhorn iteration method~\cite{cuturi2013sinkhorn}. 
Furthermore, we propose a novel semi-debiased Sinkhorn loss to optimally control the bias-variance trade-off when estimating gradients of this OT distance in the privacy preserving setting.
Since our approach does not rely on adversarial components, it avoids any training instabilities and removes the need for early stopping (stopping before catastrophic divergence of GANs, as done, for example, in~\cite{brock2018biggan}). This makes our method easy to train and deploy in practice. 
To the best of our knowledge, DP-Sinkhorn is the first fully OT-based approach for differentially private generative modeling.



In summary, we make the following contributions: (i) We propose DP-Sinkhorn, a flexible and robust optimal transport-based framework for training differentially private generative models. 
(ii) We demonstrate a novel technique to finely control the bias-variance trade-off of gradient estimates when using the Sinkhorn loss. 
(iii) Benefiting from these technical innovations,
we achieve state-of-the-art performance on widely used image modeling benchmarks for varying privacy budgets, both in terms of image quality (as measured by FID) and downstream image classification accuracy.
Finally, we present informative RGB images generated under strict differential privacy without the use of public data, with image quality surpassing that of concurrent works.

\input{related_works}
\input{background}
\input{method}
\input{experiments}

\input{discussion}
\vspace{-1mm}
\begin{ack}
\vspace{-1mm}
This work was funded by NVIDIA. Tianshi Cao and Alex Bie acknowledge additional revenue from Vector Scholarships in Artificial Intelligence, which are not in direct support of this work.
\end{ack}

\Urlmuskip=0mu plus 1mu\relax
\def\UrlBreaks{\do\/\do-}
\newpage
\bibliography{main}
\bibliographystyle{ieeetr}
\newpage

\newpage
\appendix

\input{appendix}
\end{document}

%% file: related_works.tex
\section{Related Works} \label{sec:related}
The task of learning generative models on private data has been tackled by many prior works. The general approach is to introduce privacy-calibrated noise into the model parameter gradients during training. A long line of works have explored combinations of GANs and differential privacy.
DPGAN~\cite{xie2018dpgan} first introduced the idea of combining differential privacy with GANs in a simple scheme where DPSGD~\cite{abadi2016diffprivacy} is used when updating the generator. This is followed up a year later by dpGAN~\cite{frigerio2019dpgan}, which adds a decaying clipping threshold that heuristically matches the decreasing gradient magnitude during training. DP-CGAN~\cite{torkzadehmahani2019dpcgan} adds class conditioning to DPGAN for generation of conditional data. PATE-GAN~\cite{jordon2018pategan} adopts the PATE framework for generative learning by using PATE~\cite{papernot2017semisupervised} to train a private student discriminator; only generated images are scored by this student discriminator to train the generator. This work is improved by G-PATE~\cite{long2019scalable}, which uses random projections and gradient quantization to directly aggregate discriminator gradients for updating the generator. Importantly, G-PATE makes the point that only the generator is released in the DP generative learning task, thus a large ($\sim$1000s) ensemble of non-private discriminators can be used to train a private generator. GS-WGAN~\cite{chen2020dpwgan} brings this idea back to DPSGD-based training, in which the gradient aggregate from an ensemble of discriminators is processed by the Gaussian mechanism. Unlike DP-GAN, this is performed on the image gradient, which has fewer dimensions than the parameter gradient. Datalens~\cite{wang2021datalens} further improves upon G-PATE by introducing TopAgg---a three step gradient compression and aggregation algorithm, which provides stable, quantized discriminator gradients at a low privacy cost.

It is well documented that GANs are unstable during training~\cite{arjovsky2017towards,mescheder2018ganconvergence} due to the non-optimality of the discriminator producing large biases in the generator gradient~\cite{bousquet2017vegan}.
This problem is critical in the context of DP, where the imposed gradient noise can increase training instabilities and interaction with private data (for example during hyperparameter tuning) should be limited. Our approach circumvents these issues by not relying on adversarial learning schemes.
Furthermore, state-of-the-art methods~\cite{chen2020dpwgan, wang2021datalens} rely on training a large number of discriminators to take advantage of the subsampling property of differential privacy. This hinders their practical usefulness as the discriminators require large amounts of GPU/TPU memory. 
In contrast, only a single generator network is trained in DP-Sinkhorn, making our approach more amenable to various hardware configurations.

Other generative models have also been studied in the DP setting. \cite{acs2018dpvae} partitions the private data in clusters and learns separate likelihood-based models for each cluster. \cite{harder2020dpmerf} uses Maximum Mean Discrepancy with random Fourier features.
While these works do not face the same stability issues as GANs, their restricted modeling capacity results in these methods mostly learning prototypes for each class. DP-Sinkhorn is better at using the modeling capacity of neural networks to produce high utility synthetic data while preserving privacy. Lastly, while \cite{takagi2020p3gm} produced strong empirical results, their privacy analysis relies on the use of Wishart noise on sample covariance matrices, which has been proven to leak privacy \cite{sarwate2017wishart}. Hence, their privacy protection is invalid in its current form.
\vspace{-3pt}

%% file: background.tex
\section{Background}
\vspace{-6pt}
\subsection{Notations and Setting} \label{sec:notations}
Let $\mathcal{X}$ denote a sample space, $\mathcal{P}(\mathcal{X})$ all possible measures on $\mathcal{X}$, and $\mathcal{Z} \subseteq \mathbb{R}^d$ the latent space. 
We are interested in training a generative model $g: \mathcal{Z} \mapsto \mathcal{X}$ such that its induced distribution $\mu = g \circ \xi$ with noise source $\xi \in \mathcal{P}(\mathcal{Z})$ is similar to observed $\nu$ through an independently sampled finite sized set of observations $D=\{\vy\}^N$. 
In our case, $g$ is a trainable parametric function with parameters $\theta$.

\subsection{Generative Learning with Optimal Transport} \label{sec:GLOT}
Optimal Transport-based generative learning considers minimizing variants of the Wasserstein distance between real and generated distributions~\cite{bousquet2017vegan,peyre2019ot}. Two key advantages of the Wasserstein distance over standard GANs, which optimize the Jensen-Shannon divergence \cite{NIPS2014_5423}, are its definiteness on distributions with non-overlapping supports, and its weak metrization of probability spaces \cite{arjovsky2017wgan}. This prevents collapse during training caused by discriminators that are overfit to training data. 

The OT framework can be formulated in either the primal or dual formulation. In WGAN and variants \cite{arjovsky2017wgan, gulrajani2017improved, miyato2018spectral}, the dual potential is approximated by an adversarially trained discriminator. These methods still encounter instabilities during training, since the non-optimality of the discriminator can produce arbitrarily large biases in the generator gradient \cite{bousquet2017vegan}. The primal formulation involves solving for the optimal transport plan---a joint distribution over the real and generated sample spaces. The distance between the two distributions is then measured as the expectation of a point-wise cost function between pairs of samples as distributed according to the transport plan. 

In general, finding the optimal transport plan is a difficult optimization problem. 
The entropy-regularized Wasserstein distance (ERWD) imposes a strongly convex regularization term on the Wasserstein distance, making the OT problem between finite samples solvable in linear time \cite{peyre2019computational}. Given a positive cost function $c: \mathcal{X} \times \mathcal{X} \mapsto \mathbb{R}^+$ and $\lambda \geq 0$, the ERWD is defined as: 
\begin{equation}
W_{c,\lambda}(\mu, \nu) = \min_{\pi \in \Pi} \int c(\vx,\vy)d\pi(\vx,\vy) + \lambda \int \log\left( \frac{d\pi(\vx,\vy)}{d\mu(\vx)d\nu(\vy)}\right)d\pi(\vx,\vy) \label{ERWD0}
\end{equation}
where $\Pi = \{ \pi(\vx,\vy) \in \mathcal{P}(\mathcal{X} \times \mathcal{X})| \int \pi(\vx, \cdot) d \vx = \nu, \int \pi(\cdot, \vy) d \vy  = \mu\}$.

The Sinkhorn divergence 
uses auto-correlation terms to reduce the entropic bias introduced by ERWD with respect to the exact Wasserstein distance, canceling it out completely for $\mu=\nu$ (\textit{i.e.} $S_{c,\lambda} (\mu, \nu)=0$ for matching $\mu=\nu$).
This results in faithful matching between the generator and real distributions. Here, we use the Sinkhorn divergence as defined in \cite{feydy2019interpolating}. 
\begin{definition} The Sinkhorn divergence between measures $\mu$ and $\nu$ is defined as:
\begin{equation}
    S_{c,\lambda} (\mu, \nu) = 2 W_{c,\lambda}(\mu, \nu) - W_{c,\lambda}(\mu, \mu) - W_{c,\lambda}(\nu, \nu) \label{sinkhron_loss}
\end{equation}
\end{definition}

\subsection{Differential Privacy} \label{sec:dpintro}
The current gold standard for measuring the privacy risk of data-releasing programs is the notion of differential privacy (DP)~\cite{dwork2006calibrating}. 
Informally, DP measures to what degree a program's output can deviate between adjacent input datasets $d$ and $d'$---sets differing by one entry. 
For a user contributing their data, this translates to a guarantee on how much an adversary could learn about them from observing the program's output.
Here, we are learning a generative model of images, while conditioning on available semantic labels. Hence, we are interested in the domain of image-and-label datasets where each image and its label constitute an entry.

\begin{figure*}
    \centering
    \includegraphics[width=1.0\linewidth]{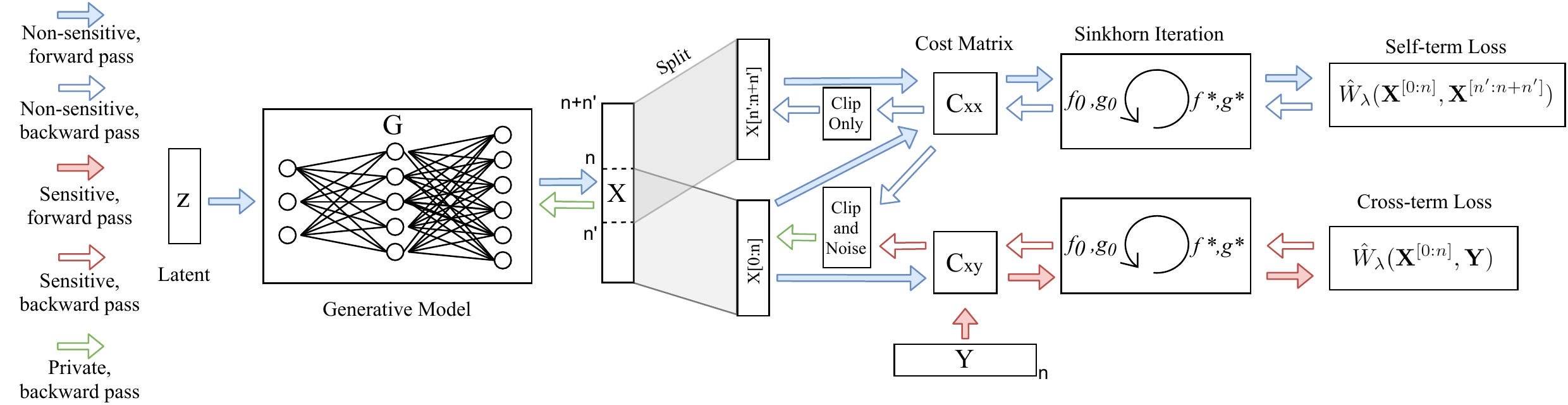}
    \caption{Flow diagram of DP-Sinkhorn for a single training iteration: Batch of generated data is split for the calculation of the cross term and the self term losses. Element-wise differences are captured in the cost matrix. Then, the losses are calculated using the Sinkhorn algorithm. In the backward pass, we impose a privacy barrier behind the generator by clipping and adding noise to the gradients at the generated image level, similar to~\cite{chen2020dpwgan}.}
    \label{fig:method_label}
    \vspace{-10pt}
\end{figure*}

A well-studied formulation of privacy, which allows tight composition of multiple queries and can be easily converted to standard definitions of DP, is provided by R\'enyi Differential Privacy (RDP)~\cite{mironov2017renyi}:

\begin{definition}(R\'enyi Differential Privacy) A randomized mechanism  $\mathcal{M}: \mathcal{D} \to \mathcal{R}$ with domain $\mathcal{D}$ and range $\mathcal{R}$ satisfies $(\alpha, \epsilon)$-RDP if for any adjacent $d, d' \in \mathcal{D}$:
\begin{equation}
    D_\alpha(\mathcal{M}(d)|\mathcal{M}(d')) \leq \epsilon,
\end{equation}
where 
$D_\alpha$ is the R\'enyi divergence of order $\alpha$. 
Also, any $\mathcal{M}$ that satisfies $(\alpha,\epsilon)$-RDP also satisfies $(\epsilon+\frac{\log {1/\delta}}{\alpha-1}, \delta)$-DP
\end{definition}
Here, $\mathcal{M}$ is a DP-learning algorithm, $d$ a training set, and $\mathcal{M}(d)$ a generator trained on $d$.
The randomized mechanism can often be dissected into a deterministic function of the dataset $\mathcal{Q}:\mathcal{D} \to \mathcal{R}$, the query, and a noise injecting random function $\mathcal{M}':\mathcal{R} \to \mathcal{R}$, the privacy mechanism, such that $\mathcal{M}(d) =\mathcal{M}'\circ \mathcal{Q}(d)$.
The sensitivity of a query $S(\mathcal{Q})$ is a property that represents the maximum magnitude of change between outputs of the query when applied to adjacent datasets.
For a query output $q=\mathcal{Q}(d)$ with sensitivity $S(\mathcal{Q})$ and standard deviation of Gaussian noise $\sigma$, the Gaussian mechanism $M(q) = q + z, \, z \sim N(0,\sigma^2)$
satisfies $(\alpha, \alpha S(\mathcal{Q})^2/(2\sigma^2))$-RDP~\cite{mironov2017renyi}. Subsampling the dataset into batches also improves privacy. The effect of subsampling on the Gaussian mechanism under RDP has been studied in \cite{wang2019subsampled, balle2018privacy, zhu2019poission}. Furthermore, privacy analysis of a gradient-based learning algorithm entails accounting for the privacy cost of single queries (possibly with subsampling), summing up the privacy cost across all queries (i.e. training iterations in our case), and then choosing the best $\alpha$. A more thorough discussion of DP can be found in 
the Appendix.

%% file: method.tex
\setlength{\intextsep}{0pt}%
\vspace{-10pt}
\begin{wrapfigure}{R}{0.52\linewidth}
\begin{minipage}{.95\linewidth}
\vspace{-10pt}
\begin{algorithm}[H]
\small
\caption{{\textit{DP-Sinkhorn}} \\
$L$ is number of categories. $\mathcal{X}$ is sample space. $M$ is size of private data set. $backprop$ is a reverse mode auto-differentiation function that takes `out', `in' and `grad weights' as input and computes the Jacobian vector product $J_\text{in}(\text{out})\cdot \text{grad weights}$. Poisson Sample and $\hat{W}_\lambda$ (via Sinkhorn iterations) are defined in Appendix. 
}
\begin{algorithmic} \label{algo:dpsinkhorn}

\STATE {\bfseries Input:} private data set $d = \{(\vy, \mathrm{l}) \in \mathcal{X} \times \{0,...,L\}\}^M$, sampling ratio $q$, noise scale $\sigma$, clipping coefficient $\Delta$, generator $g_\theta$, learning rate $\alpha$, entropy regularization $\lambda$, debiasing resample fraction $p$, total steps $T$. 
\STATE {\bfseries Output:} $\theta$
\STATE $n = q*M$, $n' = floor(n*p)$
\FOR{$t=1$ {\bfseries to} $T$}
\STATE Sample $\my \leftarrow \text{Poisson Sample}(d, q)$, 
\STATE $\mathbf{Z} \leftarrow \{\mathbf{z}_i\}_{i=1}^{(n+n')} \overset{i.i.d.}{\sim} \text{Unif}(0,1)$
\STATE $L_x \leftarrow \{ \mathrm{l}_i\}_{i=1}^{(n+n')} \overset{i.i.d.}{\sim} \text{Unif}(0,...,L)$ 
\STATE $\mx \leftarrow \{\vx_i = g_\theta(\mathbf{z}_i, \mathrm{l}_i)\}_{i=1}^{(n+n')}$
\TC{
\STATE $\mathbf{G}_\mx \leftarrow \nabla_\mx \hat{S}_{c, \lambda, p}(\mx, \my) $
\STATE $\mathbf{G}_{\mx^{[0:n]}} \leftarrow \{clip(\mathbf{G}_{\mx^i}, \Delta) + 2\Delta\sigma\mathcal{N}(\vec{0}, \mathbb{I})\}_{i=0}^{n-1}$
\STATE $\mathbf{G}_{\mx^{[n:n+n']}} \leftarrow \{clip(\mathbf{G}_{\mx^i}, \Delta)\}_{i=n}^{n+n'-1}$}
\STATE $\mathbf{G}_\theta \leftarrow backprop(\mx, \theta, \mathbf{G}_\mx)$
\STATE $\theta \leftarrow \theta - \alpha * Adam(\mathbf{G}_\theta)$
\ENDFOR
\end{algorithmic}
\end{algorithm}
\vspace{-10pt}
\end{minipage}
\end{wrapfigure}
\setlength{\intextsep}{12pt}%

\section{Differentially Private Sinkhorn}

We propose DP-Sinkhorn (Fig.~\ref{fig:method_label}), an OT-based method to learn differentially private generative models that avoids the training instability issues of GANs.
In this section, we first provide an overview of DP-Sinkhorn, followed by our novel semi-debiased Sinkhorn loss function. We then analyze the privacy protection of DP-Sinkhorn, and discuss some design considerations.

\subsection{Overview of DP-Sinkhorn}
DP-Sinkhorn aims to stably and robustly train generative models on high dimensional data (e.g. images) while preserving the privacy of training data. As discussed in Sec. \ref{sec:related}, current state-of-the-art methods in privacy-preserving data generation are reliant on adversarial training schemes that are not robust, unstable, and complicated to train. DP-Sinkhorn leverages advancements in OT-based generative learning to do away with the adversarial training scheme. 
Specifically, training a generative model with DP-Sinkhorn is a straightforward end-to-end iterative loss minimization process. In each iteration, 
data produced by the generator are split according to the debiasing ratio into a ``cross" group and a ``debiasing" group. Empirical OT distances are calculated between the ``cross" group and the real data, and between the ``debiasing" group and the ``cross group". Gradients of the OT distances with respect to the generated data are calculated and backpropagated to the generator. Privacy protection is enforced by clipping and adding noise to the gradients of the ``cross" group during backpropagation.

\subsection{Estimating Sinkhorn Divergence with Semi-Debiased Sinkhorn Loss} 
Sinkhorn divergence, as expressed in Eq.~\ref{sinkhron_loss}, involves integration over the sample space. Empirical estimation of Eq.~\ref{sinkhron_loss} based on finite samples is required to train a generative model through gradient-based optimization.
A solution suggested by previous works~\cite{feydy2019interpolating, peyre2019ot} would be to replace $\mu$ and $\nu$ with empirical samples from each distribution. 
\begin{definition} \label{def:empiricalsinkhorn}
The empirical Sinkhorn loss computed over a batch of $n$ generated examples $\mx$ and $m$ real examples $\my$ is defined as~\cite{feydy2019interpolating}:
\begin{align} \label{eq:empiricalsinkhorn}
    \hat{S}_{c, \lambda}(\mx, \my) &= 2 \hat{W}_\lambda (\mx, \my) -\hat{W}_\lambda (\mx, \mx) \nonumber \\
    & \quad 
    - \hat{W}_\lambda (\my, \my),
\end{align}
where $\hat{W}_\lambda(\ma,\mb) = C_{\ma\mb}\odot P^\lambda_{\ma\mb}$.\footnote{$\odot$ is the Hadamard product.} $C_{\ma\mb} \in \mathbb{R^{+}}^{n\times m}$ with $C_{i,j} = c(\vx_i, \vy_j)$ ($\vx_i,\vy_j$ are rows of $\ma,\mb$) is the cost matrix between $\ma$ and $\mb$, and $P^\lambda_{\ma\mb}$ is the approximate optimal transport plan that empirically minimizes $\hat{W}_\lambda(\ma,\mb)$ computed by the Sinkhorn algorithm. 
\end{definition}

However, \cite{salimans2018improving} showed that the gradients of $\hat{S}_{c, \lambda}(\mx, \my)$ are biased estimates of the gradients of $S_{c, \lambda}(\mu, \nu)$, computed over the population. Instead, they proposed a loss formulation that produces unbiased gradients using additional independently drawn samples:
\begin{definition}\label{def:debiasedsinkhorn}
Following the notations of Def.~\ref{def:empiricalsinkhorn}, let $\mx'$ and $\my'$ denote a second batch of generated and real examples. The debiased Sinkhorn loss is then defined as~\cite{salimans2018improving}:
\begin{equation}\label{eq:debiasedsinkhorn}
     \hat{S}_{c, \lambda}(\mx, \my, \mx', \my') =  2\hat{W}_\lambda (\mx, \my) - \hat{W}_\lambda (\mx, \mx') - \hat{W}_\lambda (\my, \my').
\end{equation}
\end{definition}

In comparison with Def.~\ref{def:empiricalsinkhorn}, Def.~\ref{def:debiasedsinkhorn} comes with higher variance (only $\hat{W}_\lambda (\mx, \my)$ contributes to variance in Def.~\ref{def:empiricalsinkhorn}). 
Unfortunately, privacy constraints in the DP setting prevent us from using very large batch sizes or very long training periods with low learning rates to effectively reduce variance. Hence, the variance incurred from using the unbiased estimator is more difficult to handle in the DP setup. Furthermore, Def.~\ref{def:debiasedsinkhorn} draws two batches of real data in every 
training step, thereby increasing the privacy cost of each step. Nonetheless, Def.~\ref{def:debiasedsinkhorn} is an unbiased estimator with better convergence properties. We now discuss how we overcome the above issues in DP-Sinkhorn.

First, we make the observation that the $\hat{W}_\lambda (\my, \my')$ term does not contribute to gradients of the generator. Hence, we can omit it from $\hat{S}_{c, \lambda}$.  
Next, we propose a loss formulation that interpolates between biased and unbiased Sinkhorn divergence. As observed in previous works, it can sometimes be beneficial to control bias-variance trade-offs through mixing biased and unbiased gradient estimators \cite{poole2019variational}.
Instead of completely resampling the generator for $\mx'$, we reuse some of the samples in $\mx$ when computing $\hat{W}_\lambda (\mx, \mx')$. This provides better control over the bias-variance trade-off when empirically estimating gradients.

\begin{definition}(Semi-debiased Sinkhorn loss) For a mixture fraction $p \in [0,1]$ and natural number $n$, 
$n' = floor(n \times p)$. Given $n+n'$ generated samples $\mx \in \mathcal{X}^{n+n'}$ and $m$ real samples $\my \in \mathcal{X}^{m}$, the semi-debiased Sinkhorn loss is defined as:
\begin{equation}\label{eq:semidebiased}
    \hat{S}_{c, \lambda, p}(\mx, \my) = 2 \hat{W}_\lambda (\mx^{[0:n]}, \my) -\hat{W}_\lambda (\mx^{[0:n]}, \mx^{[n':n+n']}),
\end{equation}
where $\mx^{[a:b]}$ denotes the contiguous rows of $\mx$ starting from $a$ and ending with $b-1$.
\end{definition}
When $p=1$, Eq.~\ref{eq:semidebiased} is equal to Eq.~\ref{eq:debiasedsinkhorn}, whereas when $p=0$, Eq.~\ref{eq:semidebiased} recovers Eq.~\ref{eq:empiricalsinkhorn} (ignoring the terms in Eqs.~\ref{eq:empiricalsinkhorn} and~\ref{eq:debiasedsinkhorn} that only depend on data $\my$ and are irrelevant during training). 

Algorithm 1 describes how Eq.~\ref{eq:semidebiased}
is used to train a generative model, while additionally modifying the gradient by adding noise and clipping to implement the privacy mechanism described below. 
Training of the generator proceeds by computing the gradient of the semi-debiased Sinkhorn loss with respect to $\mx$. 
Please also see the Appendix for more details.

\subsection{Privacy Protection}
Information about real data enters the generator through loss function gradients with respect to the generated images.
\TC{We achieve privacy by sanitizing the gradients of the loss function with respect to each generated image using the Gaussian mechanism.}
Let $dim(\mathcal{X})$ denote the image size,
we denote the gradients of the semi-debiased Sinkhorn loss as $\mathbf{G} \in \mathbb{R}^{(n+n')\times dim(\mathcal{X})}  = \nabla_\mx \hat{S}_{c, \lambda, p}(\mx, \my)$. 
\TC{We apply clipping and add noise to $\mathbf{G}$ to obtain a sanitized gradient $\Tilde{\mathbf{G}} \in \mathbb{R}^{(n+n')\times dim(\mathcal{X})}$ such that:
\begin{gather*}
    \begin{cases}
    \Tilde{\mathbf{G}}^{[i]} = \mathbf{G}^{[i]} \cdot \min{(\frac{\Delta}{|| \mathbf{G}^{[i]}||_2}, 1)} + \gamma \, , \, i \in \{0, \dots, n-1\} \, , \, \gamma \sim \mathcal{N}(0, \Delta^2\sigma^2)\\
    \Tilde{\mathbf{G}}^{[i]} = \mathbf{G}^{[i]} \cdot \min{(\frac{\Delta}{|| \mathbf{G}^{[i]}||_2}, 1)} \, , \, i \in \{n, \dots, n+n'-1\} 
    \end{cases} 
\end{gather*}
}
\TC{The sensitivity of $\Tilde{\mathbf{G}}^{[i]}$ is $\max_{\my, \my'}||\Tilde{\mathbf{G}}^{[i]}(\my)- \Tilde{\mathbf{G}}^{[i]}(\my')||_2 \leq 2\Delta$. 
Rows $n$ to $n+n'$ of $\mathbf{G}$ are used only for debiasing, such that $\nabla_{\mx^{[n:n+n']}}\hat{W}_\lambda (\mx^{[0:n]}, \my)=0$, i.e. $\mathbf{G}^{[n:n+n']}$ contains no information about $\my$. Therefore, noise does not need to be added to rows $n$ to $n+n'$, but we apply clipping to maintain a consistent scale of all gradient magnitudes.}
The following theorem states the privacy guarantee of DP-Sinkhorn's gradient updates, with proofs in the appendix:
\begin{theorem}\label{thm:privacy} \TC{For clipping constant $\Delta$ and noise vector $\gamma \sim \mathcal{N}(0,\Delta^2\sigma^2)$, releasing $\Tilde{\mathbf{G}}$ satisfies $(\alpha, 2 \alpha n/\sigma^2)$-RDP.}
\end{theorem}
We use the RDP accountant with Poisson subsampling proposed in \cite{zhu2019poission} for privacy composition across batches and updates. Note that the batch size of $\mx$ is kept fixed, while the batch size of $\my$ follows a binomial distribution due to Poisson subsampling.

\subsection{Design Considerations}
\paragraph{Advantages of primal form OT}
When compared to WGAN~\cite{arjovsky2017wgan}, learning with primal form OT (such as Sinkhorn divergence) has distinct differences. 
While both are approximations to the exact Wasserstein distance, the source of the approximation error differs. WGAN's source of error lies in the sub-optimality of the dual potential function. Since this potential function is parameterized by an adversarially trained deep neural network, it enjoys neither convergence guarantees nor feasibility guarantees. Furthermore, the adversarial training scheme can cause the discriminator and generator to change abruptly every iteration to counter the strategy of the other player from the previous iteration~\cite{mescheder2017numerics}, resulting in non-convergence. These challenges are exacerbated in the DP setting. In contrast, the suboptimality of the transport plan when computing Sinkhorn divergence can be controlled by using enough iterations, and the bias introduced by entropic regularization can be controlled by using small $\lambda$ values. Training with the Sinkhorn divergence does not involve any adversarial training, converges more stably, and reaps the benefits of OT metrics at covering modes.



\paragraph{Cost function} The choice of the element-wise cost $c(\vx, \vy)$ 
influences the type of images produced by the generator. We consider a mixture between pixel-wise $L_1$ and squared $L_2$ losses. $L_2$ loss has smooth gradients that scale with the difference in pixel value, whereas the gradient of $L_1$ loss is constant in magnitude for each pixel that differs. Therefore, while $L_2$ loss can quickly rein in outlier pixel values, $L_1$ loss can encourage generated image pixels to closely match those of the real image, promoting sharpness. We define the element-wise cost function as $c_m(\vx, \vy) = L_2(\vx, \vy)^2 + m\,L_1(\vx, \vy)$, 
where $L_2(\vx, \vy) = ||\vx-\vy||_2$, $L_1(\vx,\vy) = |\vx - \vy|$ and $m$ is a scalar mixture weight. Class conditioning is also achieved through the cost function by concatenating a one-hot class embedding to both the generated images and real images, similar to the approach used in \cite{salimans2018improving}. Intuitively, this works by increasing the cost between image pairs of different classes, hence shifting the weight of the transport plan ($P^*_\lambda$ in Eq.~\ref{eq:empiricalsinkhorn}) towards class-matched pairs.
\footnotetext[3]{The G-PATE~\cite{long2019scalable} authors report much more accurate classification results than reported in \cite{chen2020dpwgan}. The visual quality of samples in both papers is roughly the same.}
\footnotetext[4]{\TC{A previous version of our paper presented results using an earlier version of DP-MERF. We now run DP-MERF using code from \url{https://github.com/ParkLabML/DP-MERF/tree/master/code_balanced}.}}

%% file: experiments.tex
\vspace{-0.2cm}
\section{Experiments}\label{sec:exp}
\vspace{-2mm}
We conduct experiments on differentially private conditional image synthesis, since our focus is on generating high-dimensional data with privacy protection. We evaluate our method on both visual quality and data utility for downstream classification tasks. 
Additional experiments and analyses of the proposed semi-debiased Sinkhorn loss can be found in the Appendix. Code will be released through the project page\footnote[5]{\url{https://nv-tlabs.github.io/DP-Sinkhorn}}.

\begin{table*}[t]
  \vspace{-6pt}
  \caption{Comparison of DP image generation results on MNIST and Fashion-MNIST at $(\epsilon, \delta) = (10,10^{-5})$-DP. Results for other methods (G-PATE \cite{long2019scalable}, DP-CGAN \cite{torkzadehmahani2019dpcgan}, GS-WGAN \cite{chen2020dpwgan}) are from \cite{chen2020dpwgan}, except DP-MERF \cite{harder2020dpmerf} and Datalens \cite{wang2021datalens}. Results are averaged over 5 runs of synthetic dataset generation and classifier training.}
  \label{table:bdp}
  \centering{
  \small
  \setlength\tabcolsep{5.5pt}
  \resizebox{\columnwidth}{!}{%
  \begin{tabular}{lclccrlccr}
    \midrule
    \multicolumn{1}{l}{\multirow{3}{*}{Method}} &
    \multicolumn{1}{c}{\multirow{3}{*}{DP-$\epsilon$}} &
    \multicolumn{4}{c}{MNIST} & 
    \multicolumn{4}{c}{Fashion-MNIST} \\
     \cmidrule(r){3-6} \cmidrule(r){7-10} 
        & & \multirow{2}{*}{FID} & \multicolumn{3}{c}{Acc (\%)}  &\multicolumn{1}{c}{\multirow{2}{*}{FID}} & \multicolumn{3}{c}{Acc (\%)} \\
      \cmidrule(r){4-6} \cmidrule(r){8-10} 
        &  &  & Log Reg & MLP & CNN &   & Log Reg & MLP & CNN \\
    \midrule
   Real data  & $\infty$ & 1.6 & 92.2 & 97.5 & 99.3 & 2.5 & 84.5 & 88.2 & 90.8 \\
   Non-priv Sinkhorn ($m=1$)  & $\infty$ & 54.2 & 89.0 & 89.0 & 91.0 & 65.8 & 78.4 & 79.1 & 73.9  \\
    Non-priv Sinkhorn ($m=3$)   & $\infty$ & 43.4 & 87.7 & 87.3 & 90.6 & 63.8  & 78.4 & 78.4  & 73.3 \\
    \midrule
     G-PATE & 10& 177.2 & 26 & 25  & 51/80.9\footnotemark[3] & 205.8 & 42 & 30 & 50/69.3\footnotemark[3]  \\
     DP-CGAN   &  10 & 179.2 & 60 & 60 & 63 & 243.8 & 51  & 50  & 46    \\
     DataLens & 10 & 173.5 & N/A & N/A & 80.66 & 167.7 & N/A & N/A & 70.61 \\
     DP-MERF\footnotemark[4] & 10 & 116.3 & 79.4 & 78.3 & 82.1 & 132.6 & \textbf{75.5} & {74.5} & \textbf{75.4}\\
     GS-WGAN & 10 & 61.3 & 79 & 79 & 80 & 131.3 & 68 & 65 & 65\\
     \midrule
     DP-Sinkhorn & 10 & \textbf{48.4} & \textbf{82.8} & \textbf{82.7} & \textbf{83.2} & \textbf{128.3} & {75.1} & \textbf{74.6} & {71.1}  \\
   \bottomrule
  \end{tabular}
  }
  }
  \vspace{-12pt}
\end{table*}
\vspace{-1mm}
\subsection{Experimental Setup}
\vspace{-1mm}
\paragraph{Datasets} We use 3 image datasets: MNIST \cite{lecun1998mnist}, Fashion-MNIST \cite{xiao2017fashion}, and CelebA \cite{liu2015celeba} downsampled to 32x32 pixels. For MNIST and Fashion-MNIST, generation is conditioned on regular class labels; for CelebA we condition on gender. 
\vspace{-2mm}
\paragraph{Metrics}
In all experiments, we compute metrics against a synthetic dataset of 60k image-label pairs sampled from the model. 
For a quantitative measure of visual quality, we report 
FID~\cite{heusel2017gans}.
To measure the utility of generated data, we assess the class prediction accuracy of classifiers trained with synthetic data on the real test sets. 
We consider logistic regression, MLP, and CNN classifiers. 



\vspace{-2mm}
\paragraph{Architectures \& Hyperparameters}
We implement DP-Sinkhorn with two generator architectures. We adopt a four layer, convolutional architecture from DCGAN \cite{radford2015DCGAN} for MNIST and Fashion-MNIST experiments, and a twelve layer residual architecture from BigGAN \cite{brock2018biggan} for CelebA experiments. Class conditioning is achieved by providing a one-hot encoding of the label to the generator, and concatenating the one-hot encoding to images when calculating the element-wise cost. We set $\lambda{=}0.05$ for MNIST and Fashion-MNIST experiments, and $\lambda{=}5$ for CelebA experiments.
Complete implementation details can be found in the Appendix.

\vspace{-2mm}
\paragraph{Privacy Implementation} Our models are implemented in PyTorch. We implement the gradient sanitization mechanism by registering a backward hook to the generator output. 
MNIST and Fashion-MNIST experiments target $(10, 10^{-5})$-DP while CelebA experiments target $(10, 10^{-6})$-DP. Details are in the Appendix.
\vspace{-2mm}


\subsection{Experimental Results on Standard Benchmarks}\label{sec:benchmarks}
\begin{figure*}[t]  
\begin{minipage}[b]{0.2\linewidth}
    \small
    \centering{G-PATE~\cite{long2019scalable}} \\
    \vspace{14pt}
    \centering{DP-CGAN~\cite{torkzadehmahani2019dpcgan}} \\
    \vspace{14pt}
    \centering{DP-MERF~\cite{harder2020dpmerf}} \\
    \vspace{12pt}
    \centering{Datalens~\cite{wang2021datalens}} \\
    \vspace{13pt}
    \centering{GS-WGAN~\cite{chen2020dpwgan}} \\
    \vspace{21pt} 
    \centering{DP-Sinkhorn} \\
    \vspace{18pt}
\end{minipage}
\begin{minipage}[b]{0.7\linewidth}
    \centering
    \includegraphics[trim=0 0 0 21,clip,width=0.47\textwidth]{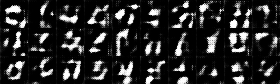}
    \includegraphics[trim=0 0 0 21,clip,width=0.47\textwidth]{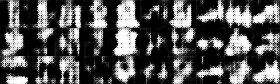}
    \includegraphics[trim=0 0 0 21,clip,width=0.47\textwidth]{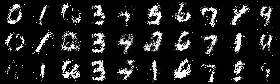}
    \includegraphics[trim=0 0 0 21,clip,width=0.47\textwidth]{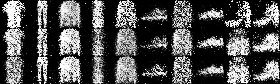}
    \includegraphics[trim=0 0 0 160,clip,width=0.47\textwidth]{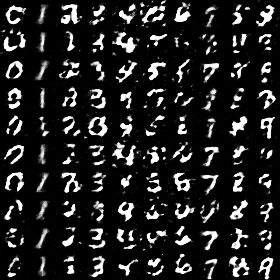}
    \includegraphics[trim=0 0 0 160,clip,width=0.47\textwidth]{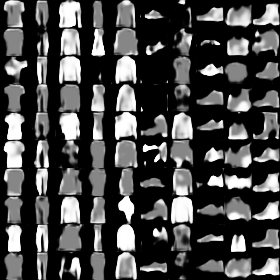}
    \includegraphics[width=0.47\textwidth]{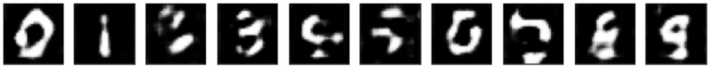}
    \includegraphics[width=0.47\textwidth]{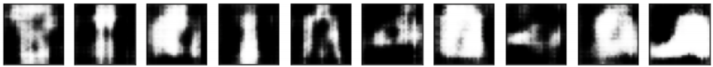}
    \includegraphics[trim=0 0 0 21,clip,width=0.47\textwidth]{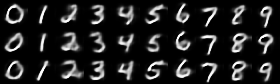}
    \includegraphics[trim=0 0 0 21,clip,width=0.47\textwidth]{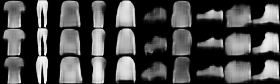}
    \includegraphics[trim=0 0 0 210px,clip,width=0.47\textwidth]{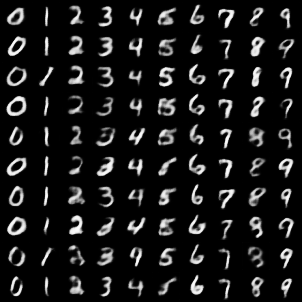}
    \includegraphics[trim=0 0 0 210px,clip,width=0.47\textwidth]{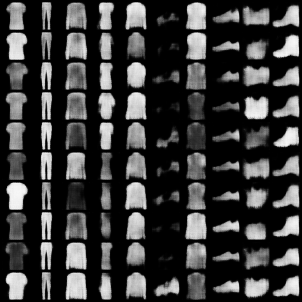}
\end{minipage}
\vspace{-5pt}
\caption{Images generated at $(10, 10^{-5})$-DP for MNIST and Fashion-MNIST by various methods. Datalens images obtained from \cite{wang2021datalens}; \TC{images of \cite{long2019scalable, torkzadehmahani2019dpcgan,chen2020dpwgan} obtained from \cite{chen2020dpwgan}.}} \label{fig:mnist_fashion_dp}
\vspace{-6pt}
\end{figure*}

\vspace{-1mm}
In Tab. \ref{table:bdp}, we compare the performance of two DP-Sinkhorn variants with other methods on MNIST and Fashion-MNIST. \TC{We use $p{=}0.4$ for the semi-debiased loss, which was determined through grid search. 
We use $m{=}1$ for MNIST and $m{=}3$ for Fashion-MNIST for weighing the $L_1$ loss in the cost function.}
\TC{On MNIST, DP-Sinkhorn generates more informative examples than previous methods, as demonstrated by the higher accuracy achieved by the downstream classifier. On the more visually complex Fashion-MNIST, DP-Sinkhorn's achieves lower FID than all baselines, while maintaining downstream accuracy similar to DP-MERF.} 
Images generated by DP-Sinkhorn are visualized in Fig.~\ref{fig:mnist_fashion_dp}. DP-Sinkhorn produces more visual diversity within each class compared to GS-WGAN, which likely benefits DP-Sinkhorn's downstream classification performance. \TC{In comparison to DP-MERF, images generated by DP-Sinkhorn carry greater resemblance to the real images, which corroborates our lower FID score.} 
\begin{wraptable}{R}{0.47\textwidth}
\vspace{-19pt}
\caption{Ablating loss functions, debiasing, and gradient perturbation mechanism on MNIST.}
 \centering
\resizebox{0.95\linewidth}{!}{
\small
\begin{tabular}{cccrrr}
\toprule
 \multirow{2}{*}{\shortstack{Image Gradient \\ Perturbation}} &
\multirow{2}{*}{Loss} & \multirow{2}{*}{Debiasing} & \multicolumn{1}{c}{\multirow{2}{*}{FID}} & \multicolumn{2}{c}{Acc (\%)} \\
\cmidrule(r){5-6}
 & & & & MLP & CNN \\
\midrule
  No & $L_2$ & No & 218.6 & 79.9 & 80.7 \\
  Yes & $L_2$ & No & 124.3 & 82.0 & 80.8\\
  Yes & $L_1$ & No & 73.9 & 68.4 & 65.7\\
  Yes & $L_1$+$L_2$ & No & 89.1 & 80.8 & 77.6\\
  Yes & $L_1$+$L_2$ & Full & 86.0 & 68.7 & 66.7 \\
  Yes & $L_1$+$L_2$ & Semi & 48.4 & 82.7 & 83.2 \\
\bottomrule
\end{tabular}}
\label{tab:ablation1}


\vspace{5pt}
\caption{DP image generation results on downsampled CelebA. We include results from \cite{wang2021datalens} for context, but note that their experiment uses a 64x64 resolution and a larger $\delta$ of $10^{-5}$.}
\resizebox{0.95\linewidth}{!}{
\begin{tabular}{lcrrr}
\toprule
\multirow{2}{*}{Method} & \multirow{2}{*}{$(\epsilon, 10^{-6})$-DP} & \multicolumn{1}{c}{\multirow{2}{*}{FID}} & \multicolumn{2}{c}{Acc (\%)} \\
\cmidrule(r){4-5}
&  &   & MLP & CNN \\
\midrule
Real data & $\infty$  & 1.1 &  91.9 & 95.0 \\
Sinkhorn  & $\infty$  & 129.5 & 80.8 & 82.2 \\
\midrule
 DP-MERF & 10 & 274.0 & 64.8 & 65 \\
 DP-Sinkhorn & 10  & \textbf{189.5} &  \textbf{76.2}& \textbf{76.3}  \\
 \midrule
 \midrule
 DataLens \cite{wang2021datalens}  & $(10, 10^{-5})$ & 320.8 & N/A & 72.9\\
\bottomrule
\end{tabular}}
\label{tab:celeba}
\vspace{-4mm}
\end{wraptable}
    

\begin{figure}
    \centering
    \includegraphics[width=0.49\textwidth]{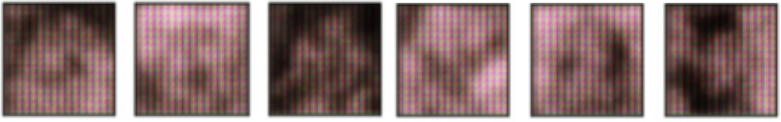}
    \includegraphics[trim=0 23 160 0,clip,width=0.24\textwidth]{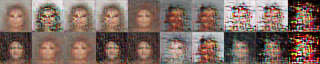}\hspace{-1pt}
    \includegraphics[trim=160 23 0 0,clip,width=0.24\textwidth]{figures/celeba/samples_ep20.png}
    \includegraphics[trim=5 560 2200 5,clip,width=0.49\textwidth]{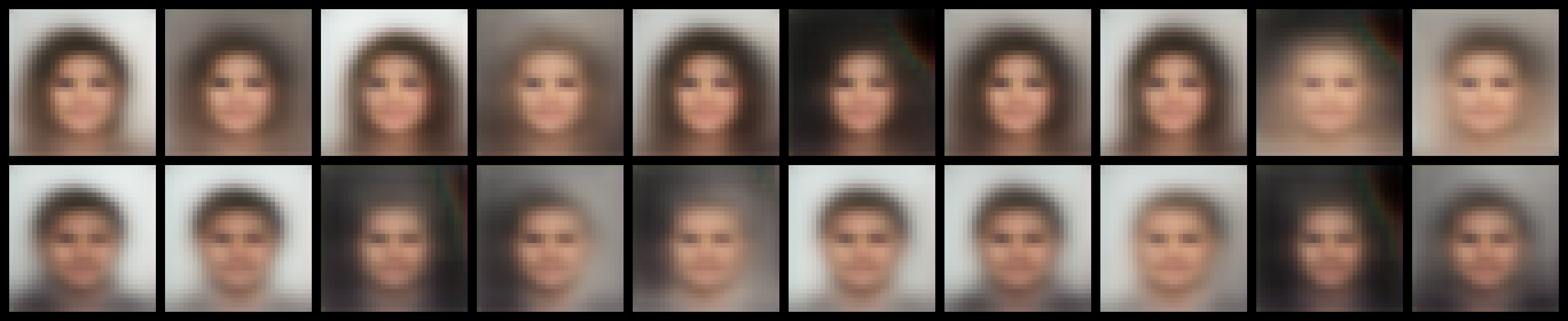}\hspace{-1pt}
    \includegraphics[trim=5 5 2200 560,clip,width=0.49\textwidth]{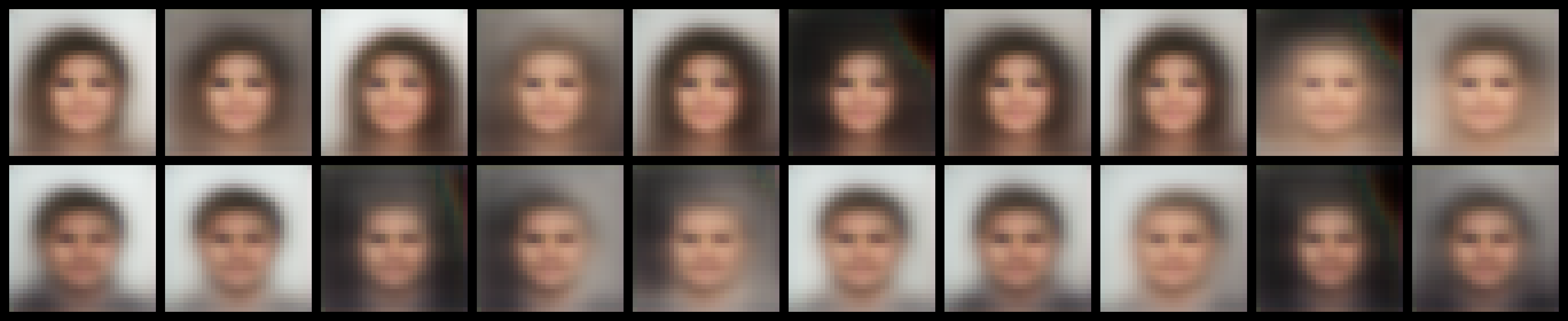}
    
    \vspace{-6pt}
    \caption{\TC{Images generated on CelebA by Datalens (top, left), DP-MERF (top, right), and DP-Sinkhorn (bottom). Datalens images obtained from \cite{wang2021datalens}.} 
    }
    \label{fig:celeba}
    \vspace{-5mm}
\end{figure}

\begin{figure}[t]
    \centering
    \begin{subfigure}[t]{0.32\textwidth}
    \centering
    \includegraphics[width=\textwidth]{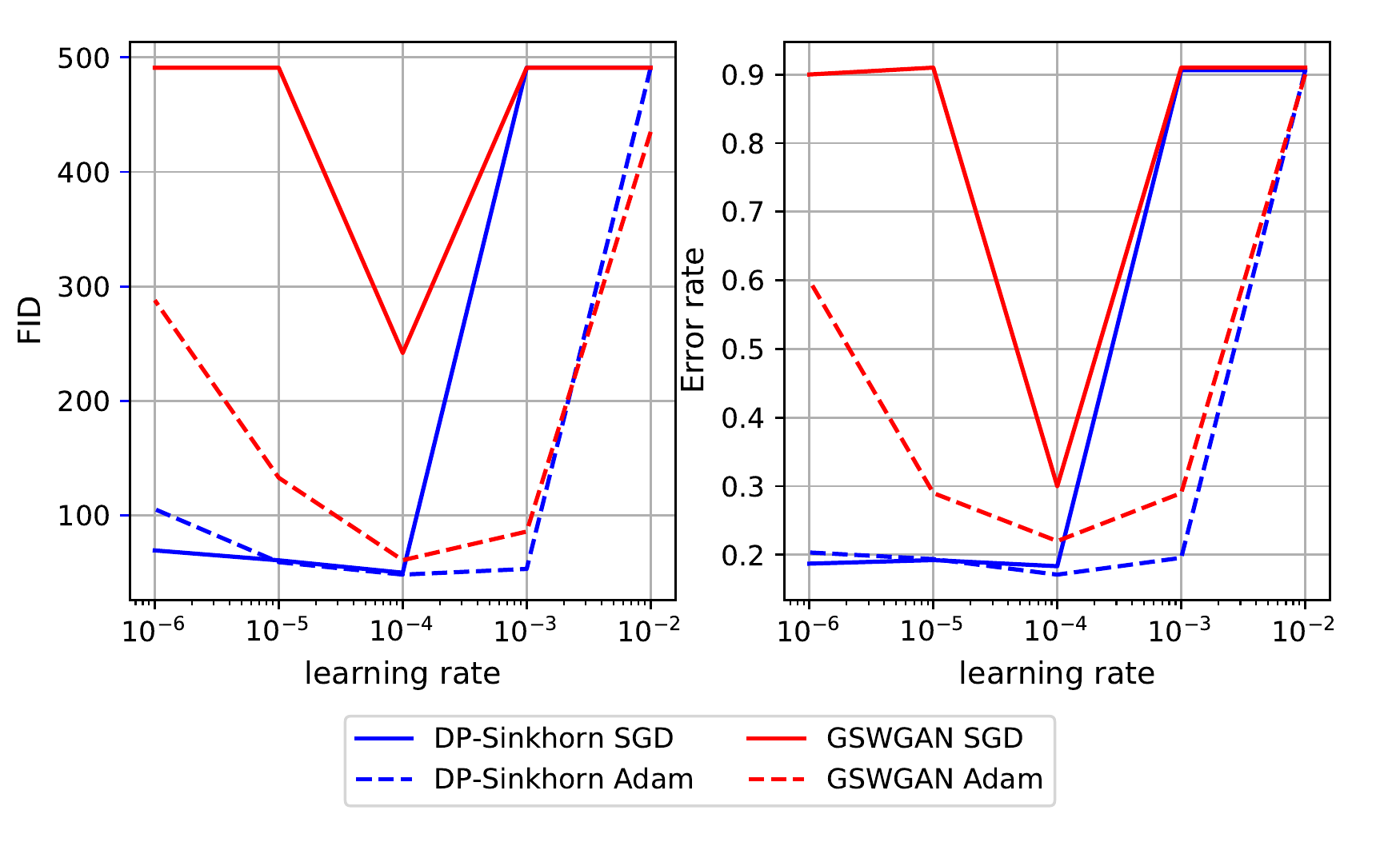}
    \caption{Comparing hyperparameter sensitivity of DP-Sinkhorn to GS-WGAN on MNIST. Error rate is calculated as $1-\textrm{Accuracy}$.} 
    \label{fig:robustness}
    \end{subfigure} \hfill
    \begin{subfigure}[t]{0.32\textwidth}
    \centering
    \includegraphics[width=\textwidth]{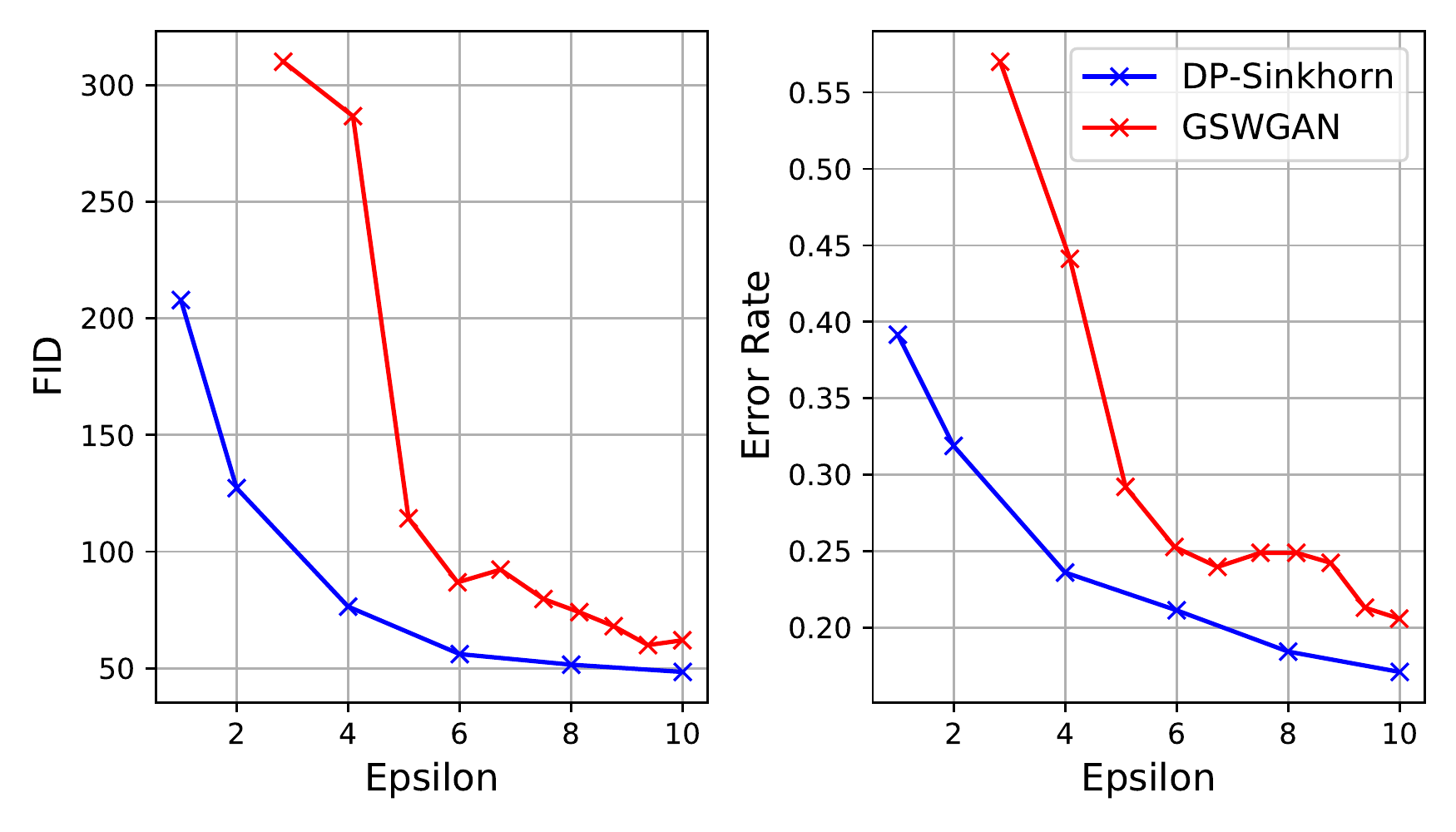}
    \caption{FID and utility of DP-Sinkhorn ($m{=}1$ and $m{=}3$), and GS-WGAN at various $\epsilon$ on MNIST. 
    }
    \label{fig:training_curve}
    \end{subfigure} \hfill
    \begin{subfigure}[t]{0.32\textwidth}
    \centering
    \includegraphics[width=\textwidth]{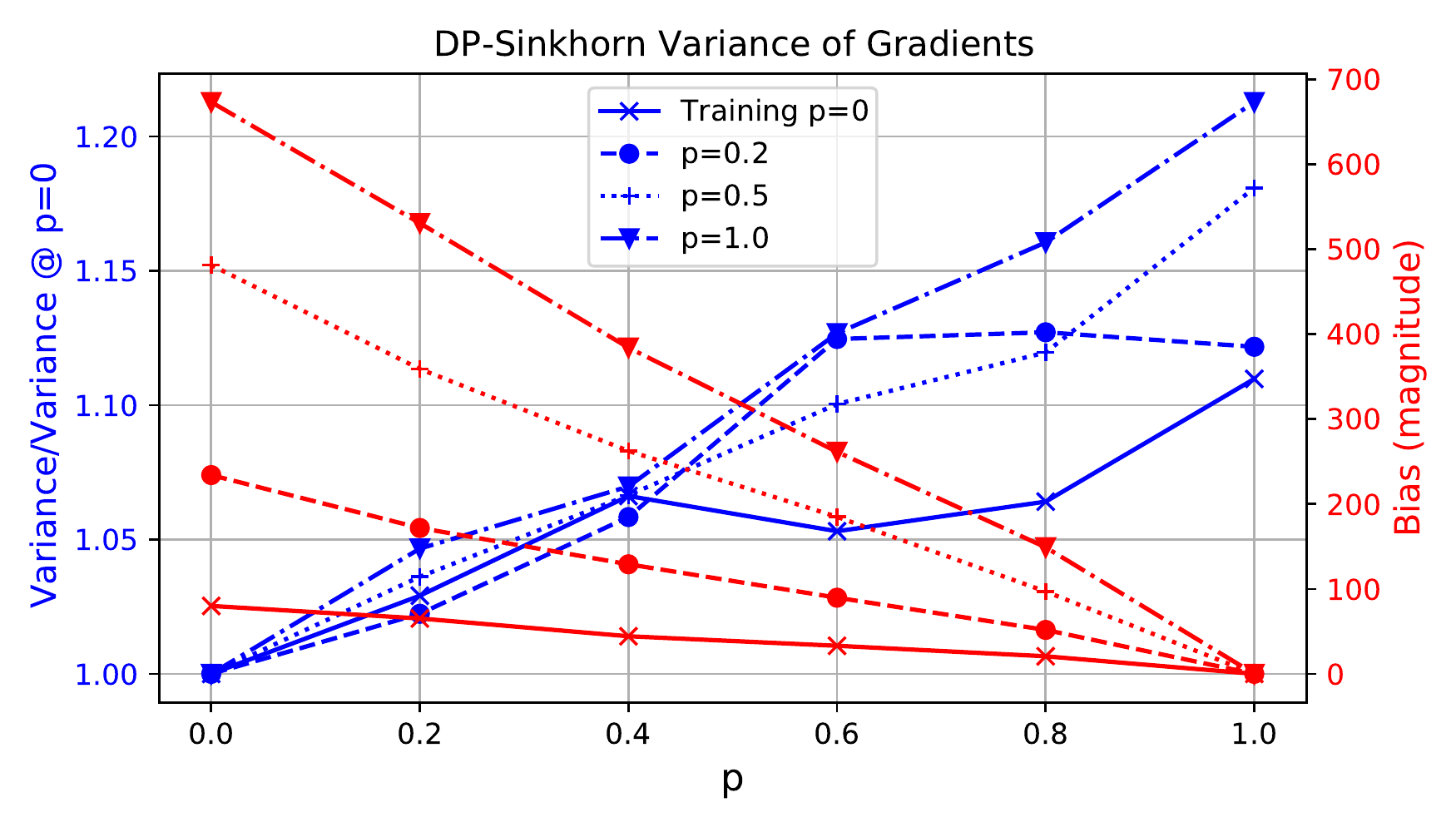}
    \caption{Bias-variance trade-off of the gradient estimator over semi-debiasing parameter $p$.}
    \label{fig:semidebias}
    \end{subfigure}
    \caption{Analyzing hyperparameter choices in DP-Sinkhorn.}
\end{figure}

\vspace{-2mm}
\paragraph{Robustness}
We evaluate the training stability of DP-Sinkhorn with different learning rates and two 
optimizers (Adam~\cite{KingmaBa2015Adam} and SGD) on MNIST. We perform the same parameter sweep on GS-WGAN for comparison\footnote[6]{\url{https://github.com/DingfanChen/GS-WGAN} (MIT License)~\cite{chen2020dpwgan}}, \TC{as it is a strong baseline that also uses gradient perturbation to enforce privacy.} Results are illustrated in Fig.~\ref{fig:robustness}.
We find that DP-Sinkhorn reliably converges for sufficiently small learning rates, and it is not sensitive to the choice of optimizer. In contrast, GS-WGAN, relying on adversarial training, suffers from non-convergence for learning rates too big or too small, and is very sensitive to the choice of optimizer. Exact numbers are reported in the Appendix. 
\vspace{-2mm}
\paragraph{Privacy Utility Trade-off} Stronger privacy protection can be attained by training DP-Sinkhorn for fewer iterations at the cost of utility and image quality. We evaluate the performance of DP-Sinkhorn at various privacy budgets and contrast it to GS-WGAN (Fig.~\ref{fig:training_curve}).
DP-Sinkhorn shows strong performance among a wide range of privacy budgets, and provides good downstream utility even at a small privacy budget of $\epsilon=2.33$, significantly outperforming GS-WGAN. 
Note that we found GS-WGAN to require significantly more memory
than DP-Sinkhorn, 
since it uses multiple discriminators for different parts of the data. In our experiments, DP-Sinkhorn can fit comfortably on an 11GB GPU, while GS-WGAN requires 24GB of GPU memory. Hence, DP-Sinkhorn is arguably more scalable to very large datsets.



\vspace{-3mm}
\paragraph{Analysis of Semi-debiased Sinkhorn Loss}
To study why our novel semi-debiased Sinkhorn loss outperforms both fully-debiased and fully-biased Sinkhorn losses, 
we evaluate bias and variance of the semi-debiased Sinkhorn loss-based gradient estimator $G_p=\nabla_\theta\hat{S}(\mx(\theta),\my)$. We sample generator gradients with respect to the semi-debiased Sinkhorn loss
with different $p$ 
and plot bias and variance (Fig.~\ref{fig:semidebias}). 
Each line represents a generator trained with a different $p$ on MNIST. For each value of $p$, we compute $G_p$ on three hundred batches of real and generated data to obtain its average and sample variance. Since $G_1$ is unbiased, we use it as the ground truth when computing bias.
Variances of each model's gradients are normalized with respect to variance of $G_0$.

We observe two prominent trends from this graph. First, as we increase $p$, bias decreases and variance increases. This requires us to find a balance in the trade-off between bias and variance. Second, we see flatter curves for generators trained with smaller $p$. As $p$ affects bias and variance through changing the number of resampled generated images, we can deduce that training with smaller $p$ likely results in greater similarity between generated images, which improves consistency across generated images at the cost of diversity. That is, if the generator is mode collapsed, $p$ would have no effect on the bias-variance trade-off, as resampling the latent variables would produce the same images. While previous works \cite{salimans2018improving} found fully-debiased $(p=1)$ Sinkhorn loss to provide higher performance, \TC{we find a moderate amount of debiasing $(p=0.4)$ to perform best}. Our hypothesis is that because training in a privacy-preserving manner is restrictive in batch size and number of iterations, the increased variance of the fully-debiased loss is more detrimental. In particular, in the DP setting we cannot simply increase batch sizes or train with more iterations and lower learning rates to counteract high loss variances, as this would incur increased privacy costs.
In contrast, our novel semi-debiasing provides control over the trade-off between consistent low-variance gradients and less biased objectives. This also demonstrates how training in the DP setting differs from the non-private setting, hence requiring new ideas and tailored methods. 
\vspace{-3mm}
\paragraph{Ablations}
We study the impact of 
perturbing image vs. parameter gradients, design of element-wise cost function, and debiasing on performance
in the MNIST benchmark. 
We start with the simplest model, using parameter gradient perturbation, $L_2$ loss and no debiasing, and incrementally add components. We \TC{use $m{=}1$ when adding $L_1$ loss, and $p{=}0.4$} when adding semi-debiasing. The clipping bound $\Delta$ is tuned separately for the variant with parameter gradient perturbation, while the other hyperparameters are kept fixed.
In Tab.~\ref{tab:ablation1}, we see that DP-Sinkhorn with parameter gradients is already competitive 
in downstream accuracy, but has poor FID in comparison to using image gradients. 
We observe that DP-Sinkhorn with $L_2$ loss yields good downsteam task performance, but has higher FID than the $L_1$ loss variant. Mixing $L_1$ and $L_2$ loss strikes a balance between better FID and downstream accuracy. We also observe that using a fully debiased gradient estimator is detrimental to performance, which we postulate is due to its high variance. The semi-debiased variant performs better than both the biased and the debiased variants. 

\vspace{-2mm}
\subsection{Experimental Results on CelebA}
\vspace{-2mm}
We also evaluate DP-Sinkhorn on downsampled CelebA. 
We evaluate whether DP-Sinkhorn is able to synthesize RGB images that are informative for downstream classification.
\TC{As baselines, we trained DP-MERF on the downsampled CelebA dataset, using the same generator architecture as ours. We also attempted to train GS-WGAN on CelebA, but couldn't obtain meaningful results after some parameter tuning.}
DP-Sinkhorn generates informative images for gender classification, as seen in Tab.~\ref{tab:celeba} (uninformative images would correspond to a ${\approx}50\%$ classification ratio).
Qualitatively, Fig. \ref{fig:celeba} shows that DP-Sinkhorn can learn meaningful representations of each semantic class (male and female) and produces some in-class variations, while avoiding details that could uniquely identify individuals.
\TC{Images generated by DP-MERF are blurrier than ours, and are less informative for the gender classification task.}
Concurrent to our work, Datalens~\cite{wang2021datalens} was also applied to gender-conditioned generation of CelebA images, albeit with a different image resolution than ours. Images generated by DP-Sinkhorn clearly resemble faces, while those generated by Datalens are blurrier.

%% file: discussion.tex
\vspace{-2mm}
\section{Conclusions}\label{sec:concl}
\vspace{-2mm}
We propose DP-Sinkhorn, a novel optimal transport-based differentially private generative model. Our approach minimizes a new semi-debiased Sinkhorn loss in a differentially private manner.
It does not require any adversarial techniques that are challenging to optimize. Consequently, DP-Sinkhorn is easy to train, which we hope will help its adoption in practice. 
We experimentally demonstrate superior performance compared to the previous state-of-the-art both in terms of image quality and on standard image classification benchmarks using data generated under DP. Our model is applicable for varying privacy budgets and is capable of synthesizing informative RGB images in a differentially private way without using additional public data. 
We conclude that 
robust models such as ours are a promising direction for differentially private generative modeling. 

\textbf{Limitations and Future Work} Our main experiments only used simple pixel-wise $L_1$- and $L_2$-losses as cost function, yet achieve better performance than GAN-based methods.
This suggests that in the DP setting complexity in model and objective are not necessarily beneficial. Nonetheless, limited image quality is the main challenge in DP generative modeling and future work includes designing more expressive generator networks that can further improve synthesis quality, while satisfying differential privacy. To this end, kernel-based cost functions may provide better performance on suitable datasets. 
Our experiments were performed on widely-used image benchmarks for differentially private generative learning. Future works may extend our method to other data types and domains. In particular, since privacy is an important consideration for medical data, applying DP-Sinkhorn to medical datasets (such as those used in~\cite{beaulieu2019privacy}) could be of high practical interest.

\textbf{Broader Impact} Our work improves the state-of-the-art in privacy-preserving generative modeling. Such advances promise significant benefits to the machine learning community, by allowing sensitive data to be shared more broadly via privacy-preserving generative models. We believe the strong performance and robustness of DP-Sinkhorn will facilitate its adoption by practitioners.
Although DP-Sinkhorn provides privacy protection in generative learning, information about individuals cannot be eliminated entirely, as no useful model can be learned under $(0,0)$-DP. This should be communicated clearly to dataset participants. 
We recognize that classifiers learned with DP can potentially underperform for minority members within the dataset~\cite{cummings_2019,kuppam_2019,agarwal_2020}, which may also be the case for classifiers trained on data produced by DP-Sinkhorn. Addressing these types of imbalances is an active area of research~\cite{grover2019bias,choi2020fair,yu2020inclusive,lee2021selfdiagnosing}.



%% file: appendix.tex
\section{Optimal Transport via the Sinkhorn Divergence} \label{appendix:ot}
In addition to the notations defined in Sec. \ref{sec:notations}, we denote the Dirac delta distribution at $\vx \in \mathcal{X}$ as $\delta_\vx$, and the standard $n$-simplex as $\mathcal{S}^n$.

Recall from Sec. \ref{sec:GLOT} that, given a positive cost function $c: \mathcal{X} \times \mathcal{X} \mapsto \mathbb{R}^+$ and $\lambda \geq 0$, the Entropy Regularized Wasserstein Distance is defined as: 
\begin{equation} \label{ERWD_appendix}
    W_{c,\lambda}(\mu, \nu) = \min_{\pi \in \Pi} \int c(\vx,\vy)d\pi(\vx,\vy) + \lambda \int \log\left( \frac{d\pi(\vx,\vy)}{d\mu(\vx)d\nu(\vy)}\right)d\pi(\vx,\vy)
\end{equation}
where $\Pi = \left\{ \pi(\vx,\vy) \in \mathcal{P}(\mathcal{X} \times \mathcal{X})| \int \pi(\vx, \cdot) d\vx = \nu, \int \pi(\cdot, \vy) d\vy  = \mu  \right\}$.

We use the Sinkhorn divergence, as defined in \cite{feydy2019interpolating}.
\begin{definition}(Sinkhorn Loss) The Sinkhorn loss between measures $\mu$ and $\nu$ is defined as:
\begin{equation}
    S_{c,\lambda} (\mu, \nu) = 2 W_{c,\lambda}(\mu, \nu) - W_{c,\lambda}(\mu, \mu) - W_{c,\lambda}(\nu, \nu)
\end{equation}
\end{definition}
For modeling data-defined distributions, as in our situation, an empirical version can be defined, too. Note that we use a slightly different notation compared to the main text, because it is more convenient to deal with empirical distributions rather than samples when relating to the dual formulation later on.
\begin{definition}(Empirical Sinkhorn loss)
The empirical Sinkhorn loss computed over a batch of $N$ generated examples and $M$ real examples is defined as:

\begin{equation} \label{eq:empiricalsinkhorn2}
    \hat{S}_{c, \lambda}(\muhat, \nuhat) = 2 C_{\mx \my} \odot P^*_{\lambda, \mx, \my} - C_{\mx \mx} \odot P^*_{\lambda, \mx, \mx} - C_{\my \my} \odot P^*_{\lambda, \my, \my}
\end{equation}
where $\muhat = \frac{1}{N} \sum_{i=1}^N \delta_{\vx_i}$, and $\nuhat = \frac{1}{M} \sum_{j=1}^M \delta_{\vy_j}$. For two samples $\ma \in \mathcal{X}^N$ and $\mb \in \mathcal{X}^M$, $C_{\ma,\mb}$ is the cost matrix between $\ma$ and $\mb$, and $P^*_{\lambda, \ma,\mb}$ is an approximate optimal transport plan that minimizes Eq. \ref{ERWD_appendix} computed over $\ma$ and $\mb$. 
\end{definition}
$P^*_\lambda$ is arrived at by iterating the dual potentials: 
\cite{cuturi2013sinkhorn} and \cite{feydy2019interpolating} have shown the following dual formulation for the discritized version of $\hat{W}_{c, \lambda}$:
\begin{equation}
    \hat{W}_{c, \lambda}(\muhat, \nuhat) = \max_{f,g \in \mathcal{S}^N \times \mathcal{S}^M } \langle \muhat, f \rangle + \langle \nuhat, g \rangle - \lambda \langle \muhat \otimes \nuhat, \exp(\frac{1}{\lambda}(f \oplus g - C_{\muhat,\nuhat})) - 1 \rangle,
\end{equation}
where $\otimes$ denotes the product measure and $\oplus$ denotes the ``outer sum'' such that the output is a matrix of the sums of pairs of elements from each vector. $C_{\muhat,\nuhat}$ is the cost matrix between each element of $\vx$ and $\vy$ who are distributed according to $\muhat$ and $\nuhat$, $C_{ij} = c(\vx_i, \vy_j)$. Then, the optimal transport plan $P^*_\lambda$ relates to the dual potentials by $P^*_\lambda = \exp(\frac{1}{\lambda}(f \oplus g - C_{\muhat, \nuhat}))(\muhat \otimes \nuhat)$. Thus, once we find the optimal $f$ and $g$, we can obtain $P^*_\lambda$ through this primal-dual relationship. We also know the first-order optimal conditions for $f$ and $g$ through the Karush-Kuhn-Tucker theorem:
\begin{equation}
        f_i = -\lambda \log \sum_{j=1}^{M} \exp(\log (\nuhat_j) + \frac{1}{\lambda}g_j - \frac{1}{\lambda} c(\vx_i, \vy_j)) \quad  g_j=-\lambda \log \sum_{i=1}^{N} \exp(\log (\muhat_i) + \frac{1}{\lambda}f_i - \frac{1}{\lambda} c(\vx_i, \vy_j))
\end{equation}
To optimize $f$ and $g$, it suffices to apply the Sinkhorn algorithm \cite{cuturi2013sinkhorn}, see Algorithm \ref{algo:sinkhorn}. Readers can refer to \cite{feydy_2020} for further details.

\section{Differential Privacy} \label{appendix:dp}

As discussed in Sec. \ref{sec:dpintro}, differential privacy is the current gold standard for measuring the privacy risk of data releasing programs. It is defined as follows~\cite{dwork2006calibrating}:

\begin{definition}(Differential Privacy)
A randomized mechanism $\mathcal{M}: \mathcal{D} \to \mathcal{R}$ with domain $\mathcal{D}$ and range $\mathcal{R}$ satisfies $(\varepsilon, \delta)$-DP if for any two adjacent inputs $d, d' \in \mathcal{D}$ differing by at most one entry, and for any subset of outputs $S \subseteq \mathcal{R}$ it holds that
\begin{equation}
    \mathbf{Pr}\left[\mathcal{M}(d) \in S \right] \leq e^{\varepsilon} \mathbf{Pr}\left[\mathcal{M}(d') \in S \right] + \delta.
\end{equation}
\end{definition}

\textbf{Gradient perturbation}:
For a parametric function $f_\params(\vx)$ parameterized by $\params$ and loss function $L(f_\params(\vx), \vy)$, usual mini-batched first-order optimizers update $\params$
using gradients $\mathbf{g}_t = \frac{1}{N}\sum_{i=1}^N \nabla_\params L(f_\params(\vx_i), \vy_i)$. 
Under gradient perturbation, the gradient $\mathbf{g}_t$ is first clipped in $L_2$ norm by constant $\Delta$, and then noise sampled from $\mathcal{N}(0,\sigma^2 \mathbb{I})$ is added.
Since differential privacy is closed under \textit{post-processing}---releasing any transformation of the output of an  $(\varepsilon,\delta)$-DP mechanism is still $(\varepsilon, \delta)$-DP \cite{dwork2014diffprivacy}---the parameters $\theta$ are also differentially private. The relation between $(\varepsilon,\delta)$ and the perturbation parameters $\Delta$ and $\sigma$ is provided by the following theorem:
\begin{theorem} \label{theorem:perturb}
 For $c^2 > 2 \log (1.25/\delta) $, Gaussian mechanism with $\sigma \geq c \Delta / \varepsilon$ satisfies $(\varepsilon, \delta)$ differential privacy. \cite{dwork2014diffprivacy}
\end{theorem}

\textbf{Subsampling}: In stochastic gradient descent (SGD) and related methods, randomly drawn batches of data are used in each training step instead of the full dataset. This subsampling of the dataset can provide amplification of privacy protection since the privacy of any record that is not in the batch is automatically protected. The ratio of sample size to population size (number of training data) is the sub-sampling ratio, commonly referred to as $q$. Smaller $q$ results in less privacy expenditure per query. 
Privacy bounds for various subsampling methods have been extensively studied and applied~\cite{dwork2006calibrating, wang2019subsampled, balle2018privacy, zhu2019poission}.

\textbf{Composition}: SGD requires the computation of the gradient to be repeated every iteration. The repeated application of privacy mechanisms on the same dataset is analyzed through \textit{composition}. Composition of the Gaussian mechanism has been first analyzed by \cite{abadi2016diffprivacy} through the moments accountant method. 

We utilize the often used R\'enyi Differential Privacy~\cite{mironov2017renyi} (RDP), which is defined through the R\'enyi divergence between mechanism outputs on adjacent datasets:

\begin{definition}(R\'enyi Differential Privacy) A randomized mechanism  $\mathcal{M}: \mathcal{D} \to \mathcal{R}$ with domain $\mathcal{D}$ and range $\mathcal{R}$ satisfies $(\alpha, \varepsilon)$-RDP if for any adjacent $d, d' \in \mathcal{D}$ it holds that
\begin{equation}
    D_\alpha(\mathcal{M}(d)|\mathcal{M}(d')) \leq \varepsilon,
\end{equation}
where 
$D_\alpha$ is the R\'enyi divergence of order $\alpha$. 
Also, any $\mathcal{M}$ that satisfies $(\alpha,\varepsilon)$-RDP also satisfies $(\varepsilon+\frac{\log {1/\delta}}{\alpha-1}, \delta)$-DP. \label{def:renyi}
\end{definition}

As discussed in the main text, RDP is a well-studied formulation of privacy that allows tight composition of multiple queries---training iterations in our case---and can be easily converted to standard definitions of DP with definition \ref{def:renyi}. Recall that for sensitivity $S$ and standard deviation of Gaussian noise $\sigma$, the Gaussian mechanism satisfies $(\alpha, \alpha S^2/(2\sigma^2))$-RDP~\cite{mironov2017renyi}. Privacy analysis of a gradient-based learning algorithm entails accounting for the privacy cost of single queries, which corresponds to training iterations in our case, possibly with subsampling due to mini-batched training. The total privacy cost is obtained by summing up the privacy cost across all queries or training steps, and then choosing the best $\alpha$.

For completeness, the R\'enyi divergence is defined as: $D_\alpha(P|Q) = \frac{1}{\alpha} \log {\EXP{\vx \in Q}{\frac{P(\vx)}{Q(\vx)}}^\alpha}$. 

\subsection{Proof of Theorem \ref{thm:privacy}}


Recalling definitions from the main text:
\TC{\begin{gather*}
\Tilde{\mathbf{G}} \in \mathbb{R}^{(n+n')\times dim(\mathcal{X})} \, \text{such that: }\\
    \begin{cases}
    \Tilde{\mathbf{G}}^{[i]} = \mathbf{G}^{[i]} \cdot \min{(\frac{\Delta}{|| \mathbf{G}^{[i]}||_2}, 1)} + \gamma \, , \, i \in \{0, \dots, n-1\} \, , \, \gamma \sim \mathcal{N}(0, \Delta^2\sigma^2)\\
    \Tilde{\mathbf{G}}^{[i]} = \mathbf{G}^{[i]} \cdot \min{(\frac{\Delta}{|| \mathbf{G}^{[i]}||_2}, 1)} \, , \, i \in \{n, \dots, n+n'-1\} 
    \end{cases} 
\end{gather*}}

\begin{theorem_r} For clipping constant $\Delta$ and noise vector $\gamma \sim \mathcal{N}(0,\Delta^2\sigma^2)$, releasing $\Tilde{\mathbf{G}}$ satisfies $(\alpha, 2 \alpha n /\sigma^2)$-RDP.
\begin{proof} The proof relies on three simple steps: \textit{(i)} Deriving the $(\alpha, \epsilon)$-RDP privacy protection for releasing \TC{$\Tilde{\mathbf{G}}^{[i]}$ for $i \in \{0, \dots, n-1\}$, following standard methods}. \textit{(ii)} Showing that $\Tilde{\mathbf{G}}^{[n:n+n']}$ carries no information about the sensitive data $\my$. \textit{(iii)} \TC{Showing that the composed gradient $\Tilde{\mathbf{G}}$ has privacy protection $(\alpha, n \epsilon)$-RDP through composition and post-processing properties.}

\textit{(i)} \TC{ Let $\Bar{\mathbf{G}}^{[i]} = \mathbf{G}^{[i]} \cdot \min{(\frac{\Delta}{|| \mathbf{G}^{[i]}||_2}, 1)}$. 
Clearly, $\max{||\Bar{\mathbf{G}}^{[i]}||_2} \leq \Delta$. Hence, the sensitivity of $\Bar{\mathbf{G}}^{[i]}$ is $\max_{\my, \my'}||\Bar{\mathbf{G}}^{[i]}(\my)- \Bar{\mathbf{G}}^{[i]}(\my')||_2 
\leq 2\Delta$.
Therefore, by standard arguments \cite{mironov2017renyi}, releasing $\Tilde{\mathbf{G}}^{[i]} = \Bar{\mathbf{G}}^{[i]} + \gamma$ satisfies $(\alpha, 2 \alpha/\sigma^2)$-RDP.}

\textit{(ii)} Further, note that $\my$ is only involved in calculating $\hat{W}_\lambda (\mx^{[0:n]}, \my)$. That is, $\hat{W}_\lambda (\mx^{[0:n]}, \mx^{[n':n+n']})$ contains no information about $\my$. 
We also have that $\nabla_{\mx^{[n:n+n]'}}\hat{W}_\lambda (\mx^{[0:n]}, \my) = 0$. Therefore, we can show that $\mathbf{G}^{[n:n+n']}$ contains no information about $\my$:
\begin{align*}
    \mathbf{G}^{[n:n+n']} &= \nabla_{\mx^{[n:n+n]'}}(2\hat{W}_\lambda (\mx^{[0:n]}, \my) - \hat{W}_\lambda (\mx^{[0:n]}, \mx^{[n':n+n'})) \\
    &= 2\nabla_{\mx^{[n:n+n]'}}\hat{W}_\lambda (\mx^{[0:n]}, \my) - \nabla_{\mx^{[n:n+n]'}}\hat{W}_\lambda (\mx^{[0:n]}, \mx^{[n':n+n']}) \\
    &= 0 -\nabla_{\mx^{[n:n+n]'}}\hat{W}_\lambda (\mx^{[0:n]}, \mx^{[n':n+n']}), \\
    &\nabla_{\mx^{[n:n+n]'}}\hat{W}_\lambda (\mx^{[0:n]}, \mx^{[n':n+n']})\quad \text{is not a function of $\my$}
\end{align*}
Consequently, $\Tilde{\mathbf{G}}^{[n:n+n']} = \TC{\{\mathbf{G}^{[i]} \cdot \min{(\frac{\Delta}{|| \mathbf{G}^{[i]}||_2}, 1)} \}_{i=n}^{n+n'-1}}$ does not contain information about $\my$.

\textit{(iii)}
\TC{The composition property of the Gaussian mechanism in RDP~\cite{mironov2017renyi} states that the $n$-fold composition of $(\alpha, \epsilon)$-RDP queries satisfies $(\alpha, n \epsilon)$-RDP. Hence, releasing $\Tilde{\mathbf{G}}^{[0:n]}$ satisfies $(\alpha, n \epsilon)$-RDP. Finally, by the post-processing property of RDP, releasing $\Tilde{\mathbf{G}} = \text{concat}(\Tilde{\mathbf{G}}^{[0:n]}, \Tilde{\mathbf{G}}^{[n:n+n']})$ enjoys the same privacy protection as $\Tilde{\mathbf{G}}^{[0:n]}$.}
\end{proof}

\end{theorem_r}

\section{Algorithms}\label{appendix:algo}
\begin{minipage}{.45\linewidth}
\begin{algorithm}[H]
\caption{Poisson Sample} 
\begin{algorithmic}
\small
\STATE {\bfseries Input }: $d = \{(\vy, \mathrm{l}) \in \mathcal{X} \times \{0,...,L\}\}^M$, \\ sampling ratio $q$
\STATE {\bfseries Output}: $\my = \{ (\vy_j, \mathrm{l}_j) \in \mathcal{X} \times \{0,...,L\} \}_{j=1}^m$, \\ $m \geq 0$
\STATE $\mathrm{s} = \{\sigma_i\}_{i=1}^M \overset{i.i.d.}{\sim} \text{Bernoulli}(q)$
\STATE $\my = \{ d_j | \mathrm{s}_j = 1 \}_{j=1}^m $
\end{algorithmic}
\end{algorithm}
\end{minipage}
\hfill
\begin{minipage}{.45\linewidth}
\begin{algorithm}[H]
\caption{Sinkhorn Algorithm $\hat{W}_\lambda (\mx, \my)$ \\ }
\begin{algorithmic} \label{algo:sinkhorn}
\small
\STATE {\bfseries Input:} $\mx=\{\vx\}^n, \my=\{\vy\}^m, \lambda$
\STATE {\bfseries Output:} $W_\lambda$
\STATE $\forall (i, j), C_{[i,j]} = c(\mx_i, \my_j)$
\STATE $\mathbf{f}, \mathbf{g} \leftarrow \vec{0}$
\STATE $\muhat, \nuhat \leftarrow \text{Unif}(n), \text{Unif}(m)$
\WHILE{not converged}
    \STATE $\forall i, \mathbf{f}_i \leftarrow -\lambda \log \sum_{k=1}^m \exp(\log(\nuhat_k) + \frac{1}{\lambda}\mathbf{g}_k - \frac{1}{\lambda} C_{[i,k]})$
    \STATE $\forall j, \mathbf{g}_j \leftarrow -\lambda \log \sum_{k=1}^n \exp(\log(\muhat_k) + \frac{1}{\lambda}\mathbf{f}_k - \frac{1}{\lambda} C_{[k,j]})$
\ENDWHILE
\STATE  $W_\lambda = \langle \muhat, \mathbf{f}\rangle +  \langle \nuhat, \mathbf{g}\rangle$
\end{algorithmic}
\end{algorithm}
\end{minipage}
\hfill

\section{Experiment Details}

\subsection{Datasets}

MNIST and Fashion-MNIST both consist of 28x28 grayscale images, partitioned into 60k training images and 10k test images. 
The 10 labels of the original classification task correspond to digit/object class. For calculating FID scores, we repeat the channel dimension 3 times.
CelebA is composed of $\sim$200k colour images of celebrity faces tagged with 40 binary attributes. We downsample all images to 32x32, and use all 162,770 training images for training and all 19,962 test images for evaluation. Generation is conditioned on the gender attribute. 
We compute FID scores between our synthetically generated datasets of size 60k and the full test data (either 10k or 19,962 images).
The MNIST dataset is made available under the terms of the Creative Commons Attribution-Share Alike 3.0 license.
The Fashion-MNIST dataset is made available under the terms of the MIT license. MNIST and Fashion-MNIST do not contain personally identifiable information.
The CelebA dataset is available for non-commercial research purposes only. It contains images of celebrity faces that are identifiable. 
\subsection{Classifiers}

For logistic regression, we use scikit-learn's implementation, using the L-BFGS solver and capping the maximum number of iterations at 5000. 
The MLP and CNN are implemented in PyTorch. The MLP has one hidden layer with 100 units and a ReLU activation. The CNN has two hidden layers with 32 and 64 filters, and uses ReLU activations. We train the CNN with dropout $(p=0.5)$ between all intermediate layers. Both the MLP and CNN are trained with Adam with default parameters while using 10\% of the training data as hold-out for early stopping. Training stops after no improvement is seen in hold-out accuracy for 30 consecutive epochs.

\begin{figure}[ht!]
    \centering
    \includegraphics[width=0.8\textwidth]{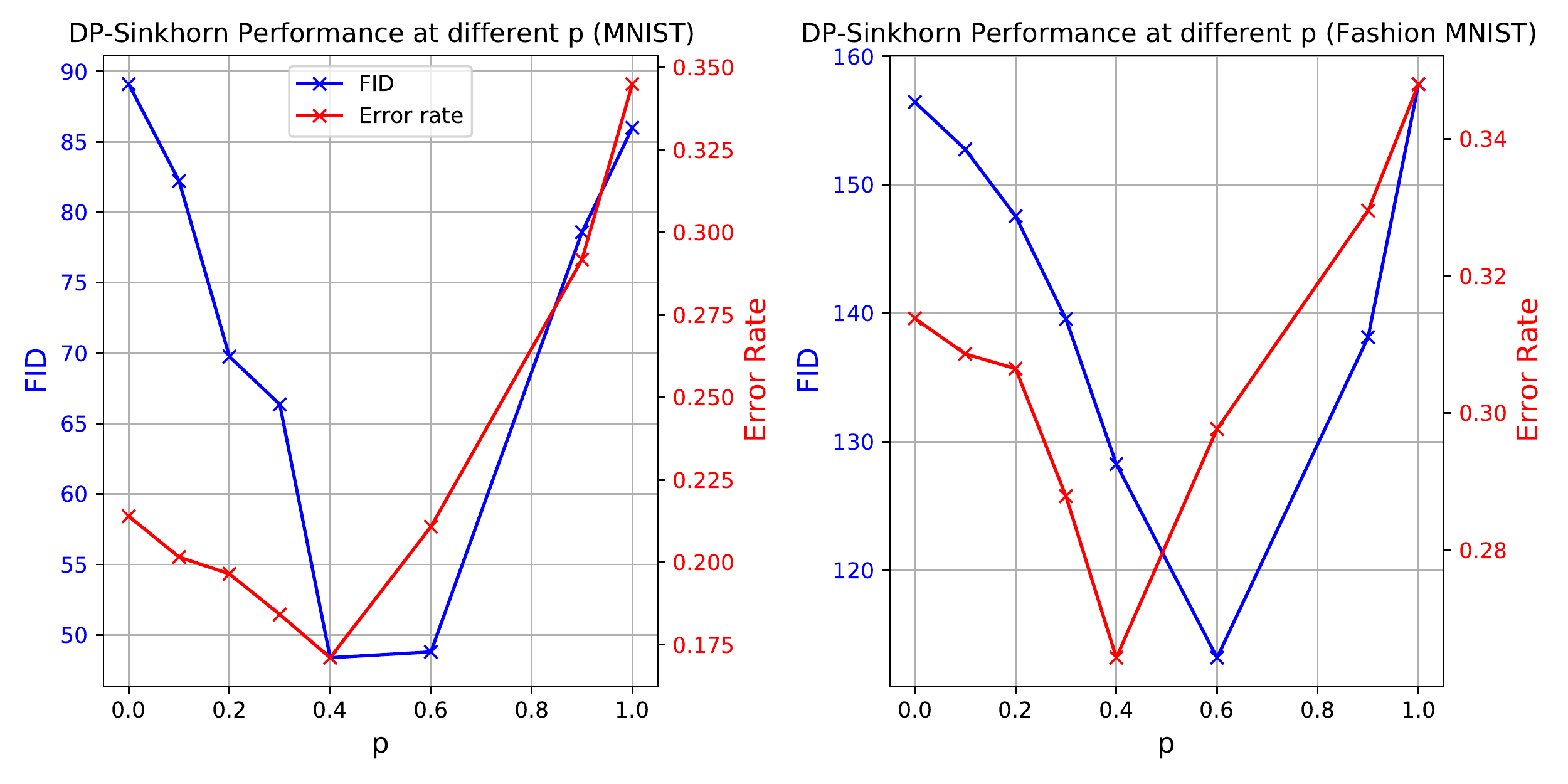}
    \caption{Effect of $p$ on DP-Sinkhorn performance. Left: performance on MNIST. Right: performance on Fashion MNIST. The performance is reported in terms of image quality (FID) and utility (error rate).}
    \label{fig:p_ablate}
\end{figure}

\subsection{Architecture, Hyperparameters, and Implementation} \label{appendix:architecture}
Our DCGAN-based architecture uses 4 transposed convolutional layers with ReLU activations at the hidden layers and tanh activation at the output layer. A latent dimension of 12 and class embedding dimension of 4 is used for MNIST and Fashion-MNIST experiments. CelebA experiments use a latent dimension of 32 and embedding dimension of 4. The latent and class embeddings are concatenated and then fed to the convolutional stack. The first transposed convolutional layer projects the input to $256 \times 7 \times 7$, with no padding. Layers 2, 3, and 4 have output depth $[128, 64, 1]$, kernel size $[4,4,3]$, stride $[2,2,1]$, and padding $[1,1,1]$. 

Our BigGAN-based architecture uses 4 residual blocks of depth 256, and a latent dimension of 32. Each residual block consists of three convolutional layers with ReLU activations and spectral normalization between each layer. Please refer to \cite{brock2018biggan} for more implementation details. Our implementation is based on \url{https://github.com/ajbrock/BigGAN-PyTorch}.

\TC{For the semi-debiased Sinkhorn loss, we set $p=0.4$ for the results reported in Tab.~\ref{table:bdp} and Tab.~\ref{tab:ablation1}. For mixing $L_1$ with $L_2$ loss, we used $m=1$ on MNIST and $m=3$ on FashionMNIST.} Hyperparameter tuning results are reported in Table~\ref{tab:hyperparam_mnist} and Table~\ref{tab:hyperparam_fashion} and visualized in Figure~\ref{fig:p_ablate}. Setting $p=0.4$ provides the best overall performance in terms of error rate and FID for both datasets.


To conditionally generate images given a target class $l$, we inject class information to both the generator and the Sinkhorn loss during training.  For the loss function, we follow \cite{salimans2018improving} and concatenate a scaled one-hot class encoding of class label $l$ to both the generated images and real images. Intuitively, this works by increasing the cost between image pairs of different classes, hence shifting the weight of the transport plan ($P^*_\lambda$ in Eq.~\ref{eq:empiricalsinkhorn}) towards class-matched pairs. A scaling constant $\alpha_c$ 
determines the importance of class similarity relative to image similarity in determining the transport plan. Thus, $\vx$ and $\vy$ are replaced by $[\vx,  \alpha_c * \mathrm{onehot}(l_x)]$ and $[\vy, \alpha_c * \mathrm{onehot}(l_y)]$ for class-conditional generation when calculating the element-wise cost. 

\TC{Hyperparameters of the Sinkhorn loss used were: $\alpha_c = 15$, and entropy regularization $\lambda = 0.05$ in MNIST and Fashion-MNIST experiments. $\lambda = 5$ is used for CelebA experiments. We use the implementation publically available at \url{https://www.kernel-operations.io/geomloss/index.html} and all other hyperparameters are kept at their default values.
For all experiments, we use the Adam \cite{KingmaBa2015Adam} optimizer. On MNIST, we set the learning rate to $10^{-4}$; on Fashion-MNIST and CelebA, we use learning rate $10^{-5}$. Other optimizer hyperparameters were left at the PyTorch default values of $\beta=(0.9,0.999)$, weight decay $2 \times 10^{-5}$.}
 
\subsection{Implementation of Differential Privacy}
For privacy accounting, we use the implementation of the RDP Accountant available in Tensorflow Privacy.\footnote{\url{https://github.com/tensorflow/privacy/}} All experiments use Poisson sampling for drawing batches of real data, and are amenable to the analysis implemented in \texttt{compute\_rdp}. 

\TC{For MNIST, we use a noise scale of $\sigma = 1.5$ and a batch size of $50$ resulting in a sub-sampling ratio of $q = 1/1200$, which gives us $\sim160,000$ training iterations (batches) to reach $\varepsilon=10$ for $\delta=10^{-5}$.
For Fashion-MNIST, we use a noise scale of $\sigma = 1.9$ and the same batch size. $\varepsilon=10$ is reached in $\sim$ 280,000 iterations. For both experiments, we use a clipping norm of $0.5$.}
For the non-private runs, we use a batch size of $500$, which improves image quality and diversity. When training with DP, increasing batch size significantly increases the privacy cost per iteration, resulting in poor image quality for fixed $\varepsilon=10$. 
\TC{For the CelebA results reported in the main text, we use a noise scale of $\sigma = 1.9$ and a batch size of $50$ resulting in a sub-sampling ratio of $q=0.00038$. At $\delta=10^{-6}$, we train for 1.7 million steps to reach $\varepsilon = 10$. The clipping norm is also set to $0.5$}

\paragraph{Computational Resources} We perform experiments on an internal in-house GPU cluster, consisting of V100 NVIDIA GPUs. Each experiment is run on a single GPU with 16GB of VRAM. On MNIST and FashionMNIST, each epoch of training takes about 50 seconds to complete. Training of the generators to $(10,10^{-5})$-DP takes 40 GPU hours to complete. On CelebA, experiments with the DCGAN architecture take $\sim$75 seconds per epoch during training, which totals to 12 GPU hours per run. BigGAN experiments take $\sim$250 seconds per epoch instead, totalling to $\sim$40 GPU hours per run. 

We estimate the total amount of GPU hours used throughout this project to be $\approx$10,000 GPU hours. We assume that an average run takes around 40 GPU hours and each round of hyperparameter tuning experiments typically consists of 16 runs. From the conceptualization of the project to its current form, we performed 16 such parameter sweeps following changes to methodology, implementation, and parameter range. This totals to 10,240 GPU hours. 

\begin{table}[ht!]
  \caption{DP-Sinkhorn ($\epsilon=10$) hyperparameter search on MNIST.}
  \label{table:hparam}
  \centering{
  \small
  \setlength\tabcolsep{5.5pt}
  \begin{tabular}{lccccrrr}
    \toprule
    \multicolumn{1}{l}{\multirow{2}{*}{$\sigma$}} &
    \multicolumn{1}{l}{\multirow{2}{*}{$\delta$}} &
    \multicolumn{1}{l}{\multirow{2}{*}{$p$}} &
    \multicolumn{1}{c}{\multirow{2}{*}{$m$}} & \multicolumn{1}{c}{\multirow{2}{*}{FID}} &
    \multicolumn{3}{c}{Acc (\%)}\\
    \cmidrule(r){6-8}
     &  &  & & & Log Reg & MLP & CNN  \\
     \midrule
    1.1 & 1.0 & 0.2 & 1.0 & 75 & 81.2 & 82.6 & 79.5 \\
    1.3 & 1.0 & 0.2 & 1.0 & 80 & 81.9 & 82.4 & 80.1 \\
    1.3 & 0.3 & 0.2 & 1.0 & 68.8 &	79.8 &	81.2 &	76.9\\
    1.3 & 0.7 & 0.2 & 1.0 & 72.0 &	79.6 &	81.1 &	77.0 \\
    1.5 & 0.5 & 0.2 & 1.0 & 74.5 &	82.3 &	82.6 &	82.1 \\
    1.5 & 0.7 & 0.2 & 1.0 & 69.2 &	80.5 &	82.2 &	80.8 \\
    1.9 & 0.5 & 0.2 & 1.0 & 76.4 & 80.2 & 80.5 & 79.2 \\
    1.9 & 0.7 & 0.2 & 1.0 & 70.3 & 82.2 & 83.4 & 80.5 \\
    1.5 & 0.5 & 0.4 & 1.0 & 48.4 & 82.8 & 82.7 & 83.2 \\
  \bottomrule
  \end{tabular}
  }
  \vspace{-5pt}
  \label{tab:hyperparam_mnist}
\end{table}

\begin{table}[ht!]
  \caption{DP-Sinkhorn ($\epsilon=10$) hyperparameter search 
  on Fashion MNIST.}
  \centering{
  \small
  \setlength\tabcolsep{5.5pt}
  \begin{tabular}{lccccrrr}
    \toprule
    \multicolumn{1}{l}{\multirow{2}{*}{$\sigma$}} &
    \multicolumn{1}{l}{\multirow{2}{*}{$\delta$}} &
    \multicolumn{1}{l}{\multirow{2}{*}{$p$}} &
    \multicolumn{1}{c}{\multirow{2}{*}{$m$}} & \multicolumn{1}{c}{\multirow{2}{*}{FID}} &
    \multicolumn{3}{c}{Acc (\%)}\\
    \cmidrule(r){6-8}
     &  &  & & & Log Reg & MLP & CNN  \\
     \midrule
    1.1 & 1.0 & 0.2 & 1.0 & 140.0 & 74.0 & 74.0 & 68.0 \\
    1.3 & 0.3 & 0.2 & 1.0 & 156.6 &	74.1 &	74.5 &	65.8\\
    1.3 & 0.7 & 0.2 & 1.0 & 157.9 &	74.0 &	74.2 &	65.8 \\
    1.5 & 0.5 & 0.2 & 1.0 & 163.4 &	73.7 &	74.4 &	67.5 \\
    1.5 & 0.7 & 0.2 & 1.0 & 161.5 &	73.9 &	74.2 &	69.3 \\
    1.9 & 0.5 & 0.2 & 1.0 & 153.8 &	73.5 &	73.9 &	69.4 \\
    1.9 & 0.7 & 0.2 & 1.0 & 154.7 &	73.9 &	73.9 &	68.4 \\
    1.9 & 0.5 & 0.4 & 1.0 & 133.0 & 73.2 & 73.2 & 69.6\\
      1.9 & 0.5 & 0.4 & 3.0 & 128.3 & 75.1 &	74.6 & 71.1\\
  \bottomrule
  \end{tabular}
  }
  \vspace{-5pt}
  \label{tab:hyperparam_fashion}
\end{table}

\begin{figure}
    \centering
    \includegraphics[width=0.9\linewidth]{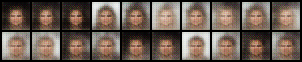}
    \includegraphics[width=0.9\linewidth]{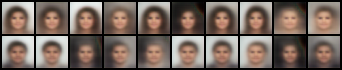}
    \caption{Additional DP-Sinkhorn generated images under $(10, 10^{-6})$differential privacy. Top two rows use DCGAN-based generator, while bottom two rows use BigGAN-based generator.}
    \label{fig:my_label}
\end{figure}

We evaluate the impact of architecture choice on the performance in the CelebA task by comparing DP-Sinkhorn+BigGAN with DP-Sinkhorn+DCGAN, under $L_2$ loss. Results are summarized in Table \ref{tab:celeba2} and visualized in Figure \ref{fig:my_label}. Qualitatively, despite reaching lower FID score, the DCGAN-based generator's images have visible artifacts that are not present in models trained with BigGAN-generators.

\begin{figure}
    \centering
    \includegraphics[width=0.45\linewidth]{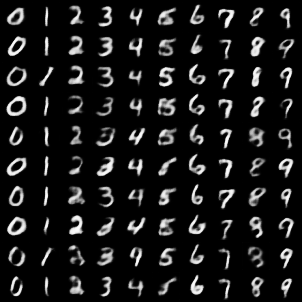}
    \includegraphics[width=0.45\linewidth]{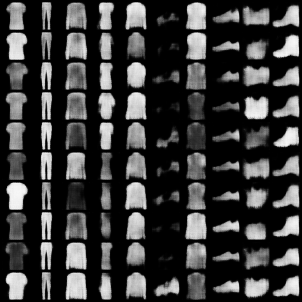}
    \caption{\TC{Additional images generated by DP-Sinkhorn, trained on MNIST (left) and Fashion-MNIST (right).}} 
    \label{fig:mnist_more_image}
\end{figure}

Additional DP-Sinkhorn samples for MNIST and Fashion-MNIST are shown in Figures \ref{fig:mnist_more_image}.
\begin{table}[h]
    \centering
    \vspace{6pt}
    \small
    \caption{Differentially private image generation results on downsampled CelebA.}
    \begin{tabular}{lcrrr}
    \toprule
    \multirow{2}{*}{Method} & \multirow{2}{*}{DP-$\epsilon$} &\multicolumn{1}{c}{\multirow{2}{*}{FID}} & \multicolumn{2}{c}{Acc (\%)} \\
    \cmidrule(r){4-5}
    &  &  & MLP & CNN \\
    \midrule
    Real data & $\infty$ & 1.1 &  91.9 & 95.0 \\
    \midrule
    DCGAN+DP-Sinkhorn & 10 & 156.7 &  74.96 & 74.62 \\
    BigGAN+DP-Sinkhorn & 10 & 168.4 &  76.18 & 75.79 \\
    \bottomrule
    \end{tabular}
    \label{tab:celeba2}
\end{table}

